%% file: main.tex
\documentclass{article} % For LaTeX2e
\usepackage{iclr2024_conference,times}

\input{math_commands.tex}

\usepackage{hyperref}
\usepackage{url}

\usepackage{graphicx}
\usepackage{subcaption}
\usepackage{etoolbox}
\usepackage{mdframed}
\usepackage{makecell}
\usepackage[ruled]{algorithm2e}

\usepackage{etoolbox}
\usepackage{xcolor}
\usepackage{multirow}
\usepackage{booktabs}

\usepackage{cleveref} % to handle multiple ref

\newbool{showcomments}
\booltrue{showcomments} % Set to true to display comments, false to hide them
\newcommand{\wang}[1]{\ifbool{showcomments}{\textcolor{brown}{[#1 - wang]}}{}}
\newcommand{\deli}[1]{\ifbool{showcomments}{\textcolor{red}{[#1 - deli]}}{}}
\newcommand{\tux}[1]{\ifbool{showcomments}{\textcolor{violet}{[#1 - tux]}}{}}

\title{Towards Codable Watermarking for Injecting Multi-Bits Information to LLMs}

\author{%
  Lean Wang\thanks{\quad Equal contribution.}\hspace{0.5em}$^{\dag,\S}$,
  Wenkai Yang$^{\ast}$$^{\ddag,\S}$,
  Deli Chen$^{\ast}$\textsuperscript{\P},
\\ \textbf{Hao Zhou}$^{\S}$\textbf{,} 
  \textbf{Yankai Lin}$^{\ddag}$\textbf{,} 
  \textbf{Fandong Meng}$^{\S}$\textbf{,} 
  \textbf{Jie Zhou}$^{\S}$\textbf{,} 
  \textbf{Xu Sun}$^{\dag}$ \\
  \textsuperscript{\dag}{National Key Laboratory for Multimedia Information Processing,} \\{\, School of Computer Science, Peking University} \\
  \textsuperscript{\ddag}{Gaoling School of Artificial Intelligence, Renmin University of China} \\
    \textsuperscript{\S}{Pattern Recognition Center, WeChat AI, Tencent Inc., China}\\
\textsuperscript{\P}{DeepSeek AI} \\
\texttt{\{lean, xusun\}@pku.edu.cn}, 
\quad\texttt{victorchen@deepseek.com}\\
\texttt{\{wenkaiyang, yankailin\}@ruc.edu.cn} \\
\texttt{\{tuxzhou, fandongmeng, withtomzhou\}@tencent.com}
}

\iclrfinalcopy % Uncomment for camera-ready version, but NOT for submission.
\begin{document}

\maketitle

\begin{abstract}
As large language models (LLMs) generate texts with increasing fluency and realism, there is a growing need to identify the source of texts to prevent the abuse of LLMs.
Text watermarking techniques have proven reliable in distinguishing whether a text is generated by LLMs by injecting hidden patterns. However, we argue that existing LLM watermarking methods are encoding-inefficient and cannot flexibly meet the diverse information encoding needs (such as encoding model version, generation time, user id, etc.).
In this work, we conduct the first systematic study on the topic of~\textbf{Codable Text Watermarking for LLMs} (CTWL) that allows text watermarks to carry multi-bit customizable information.
First of all, we study the taxonomy of LLM watermarking technologies and give a mathematical formulation for CTWL. 
Additionally, we provide a comprehensive evaluation system for CTWL: (1)~watermarking success rate, (2)~robustness against various corruptions, (3)~coding rate of payload information, (4)~encoding and decoding efficiency, (5)~impacts on the quality of the generated text.
To meet the requirements of these non-Pareto-improving metrics, we follow the most prominent vocabulary partition-based watermarking direction, and devise an advanced CTWL method named~\textbf{Balance-Marking}. 
The core idea of our method is to use a proxy language model to split the vocabulary into probability-balanced parts, thereby effectively maintaining the quality of the watermarked text.
% We take another language model to guide the vocabulary partition, leading to a probability-balanced vocabulary splitting which can effectively maintain the watermarked text quality.
% Compared to the random vocabulary partitioning, a probability-balanced vocabulary partition can significantly improve the quality of the generated text.
Extensive experimental results show that our method outperforms the baseline under comprehensive evaluation.
% We also analyzed how to select the proxy models for deploying our method in different LLM application scenarios.
Our code is available at \url{https://github.com/lancopku/codable-watermarking-for-llm}.

% We hope this work can raise the community’s awareness of the importance of CTWL and inspire further research on designing more efficient, practical, and robust watermarking methods for LLMs.
\end{abstract}

\section{Introduction}

%\footnote{*Equal contribution. Work done while Lean Wang and Wenkai Yang were in Tencent internship.}
Recently, with the explosive development of Large Language Models~(\textbf{LLMs})~\citep{chatgpt,llama}, there has been growing concern in the community about the potential negative effects of the AI-generated content (\textbf{AIGC}). For instance, LLMs could be exploited to produce fake news, encyclopedia entries, or academic papers. 
Hence, there is an urgent need to reliably distinguish between human-written and AI-generated texts. 

Text watermarking~\citep{review_of_digital_watermarking,watermarking_statistical_mt} aims to inject hidden patterns into the generated text, and detect the specific patterns to determine the source of text. 
The most representative line of LLM watermarking methods~\citep{watermark_llm, sweet,provable_robust_watermark} injects the watermark by controlling the available part of the vocabulary during LLM's decoding process and can detect whether a text contains a watermark with high accuracy and low false positive rate.
However, we argue that these existing LLM watermarking methods encode too limited information (only 1 bit of information - whether the text is generated by one specific model or not), and can not satisfy the increasing demand for customizing information in the application of LLMs (for example, embedding model and version information in the watermark can effectively trace the source of a text among multiple LLMs, etc.). 

% \begin{figure}[t]
%     \centering
% \includegraphics[width=0.95\textwidth]{figs/plot1.pdf}
%     \caption{Non-Codable LLM Watermarking VS Codable LLM Watermarking.}
%     \label{introduction_plot}
% \end{figure}

\begin{table}[t]
\centering
\caption{Taxonomy of LLM watermarking technologies, with a representative work of each direction attached. It can be found that existing LLM watermarking methods either do not make full use of the generation ability of LLMs, or lack customized watermarking information. Our work simultaneously addresses both of these issues, filling the gap in this line of academic research.}

\resizebox{.97\columnwidth}{!}
{
\begin{tabular}{c|ll}
\toprule
\multirow{2}{*}{\begin{tabular}[c]{@{}c@{}} 
\textbf{Watermark}\\ \textbf{Information}\end{tabular}} 
& \multicolumn{2}{c} {\textbf{Watermark Injection Timing}}         
\\  \cmidrule{2-3}    
& \multicolumn{1}{c}{Post-process after LLM Generation}
& \multicolumn{1}{c}{Integrate with LLM Generation}      
 \\ \midrule 
 One-Bit             
& Black-Box Watermarking~\citep{black_box_watermark}
& LLM Watermarking~\citep{watermark_llm}      
\vspace{0.06in} \\ 
Multi-Bits  
& Natual Language Watermarking~\citep{acl_muilit_bit} 
& Codable Text Watermarking for LLMs~(This Work)
\\ \bottomrule

\end{tabular}}   
\label{tab:watermark_classification}
\end{table}

\textbf{Present Work.} We conducted the first systematic study on the topic of Codable Text Watermarking for LLMs (\textbf{CTWL}), which allows the watermark injected into LLM-generated text to carry more customizable information. 
First and foremost, given the fact that the boundary and definition of CTWL remain unclear, we start with a taxonomy study on watermarking technologies~(Section~\ref{sec: related_work}) and then give a 
mathematical formalization of the CTWL problem~(Section~\ref{sec: format}). 
Then, we propose a comprehensive evaluation system for CTWL~(Appendix~\ref{sec: evaluation}), which consists of the following 5 criteria: (1)~\textbf{Success Rate}: the LLM watermarking method should have high success rates of recovering message from watermarked texts and distinguishing between human-written and watermarked ones. (2)~\textbf{Robustness}: the watermarking method should remain a high success rate when facing different challenging attacks (such as copy-paste attack, substitution attack, etc.). (3)~\textbf{Coding Rate}: the ratio of text token number to information bit number should be high. (4)~\textbf{Computational Complexity}: the computation cost of watermark writing and decoding should meet the practical hardware and latency requirements. (5)~\textbf{Text Quality}: the impact of adding complex watermark patterns on the quality of generated texts should be minimal. 

Therefore, devising a practically applicable CTWL method is a challenging task, since it is difficult to make Pareto improvements in these conflicting metrics.
% (for example, when the coding rate increases, the robustness and text quality will inevitably decrease).
% ; and when the encoding and decoding methods are sufficiently complex to ensure the success rate and robustness, the computation cost will become prohibitive. 
In this work, we first extend the random vocabulary splitting idea from~\citet{watermark_llm}, and devise the~\textbf{Vanilla-Marking} method to encode a multi-bit message in watermark. 
However, this naive baseline perform poorly in maintaining generated text quality, since multi-bit watermark is more complex than one-bit watermark, thus injecting multi-bit watermark causes a greater impact on the text quality.

% the random splitting can result in a severely unbalanced probability distribution of the available/unavailable parts (e.g. when most high-probability tokens are in the unavailable partition, the watermarking algorithm will fail). 
To overcome this challenge, and inspired by the idea that a probability-balanced vocabulary partition can more effectively ensure the quality of the watermarked text, we devise an advanced CTWL method named~\textbf{Balance-Marking}. 
% simultaneously satisfy the various evaluation dimensions mentioned above. 
Specifically, we leverage a proxy language model (\textbf{proxy-LM}) to guide the vocabulary partitioning instead of random splitting, ensuring that the probabilities of the available/unavailable parts are close to 50-50. 
This balance partition avoids the excessive impact of
watermark on text quality caused by random vocabulary splitting(e.g. most high-probability tokens are in the unavailable part). 
Also, this mechanism implicitly helps the watermark algorithm to skip low-entropy segments of text, which further ensures the text quality.
Moreover, we can achieve a flexible and customizable watermarking algorithm deployment for different LLM application scenes by switching the proxy-LM.
For example, we can take a small language model as the proxy-LM to ensure a higher inference efficiency; 
we can take a public language model as the proxy-LM and achieve an open watermarking protocol among different LLM providers without the leakage of the LLMs.~\footnote{A more detailed discussion on the application scenarios of codable text watermark and the selection of proxy language models is presented in the Appendix~\ref{sec: scenes}.} 
Extensive experimental results demonstrate that the Balance-Marking method surpasses the Vanilla-Marking method across different sizes of LLMs (OPT-1.3B~\citep{Zhang2022OPTOP}, LLaMA-7B and LLaMA-13B~\citep{llama}). 
Additional experiments are performed to assess the influence of crucial modules within our approach. Furthermore, we analyze the application scenarios of CTWL and potential avenues for future research.
% we can also use the LLM itself as the proxy-LM to ensure the generated text quality
% The proxy-LM can either be the LLM itself when the model company only wants itself to decode the watermark information, or be a smaller and public language model (e.g, GPT-2~\citep{gpt2}) with comparable performance to original LLM but has a higher inference efficiency.
% We hope that this work will draw great attention from the community to the topic of codable text watermarking, and inspire more innovative works on the development of datasets, evaluation, and implementation for CTWL.

\section{Related Work}
\label{sec: related_work}
The current studies on identifying LLM-generated texts can be mainly divided into two categories: detecting-based~(introduced in Appendix~\ref{appendix:detection_method}) and watermarking-based methods. 
The watermarking-based method has proven more effective and reliable~\cite{reliably_detect} and we focus on the watermarking-based method.
% Watermarking can be applied to any model without additional parameter training and has proven more effective than the classification-based ~\citep{gpt-2-release,distilbert,openai_classifier} or statistical-based~\citep{gptzero,detectgpt} detection methods~\citep{reliably_detect}.
% \paragraph{Waterm Methods.}
Text watermarking technology marks text by injecting human-invisible hidden patterns into the text, and then determines the text source by detecting whether the text contains a watermark~\citep{review_of_digital_watermarking,watermarking_statistical_mt,adversarial_watermarking,tracing_text_provenance}. 
We analyze the taxonomy of LLM watermarking methods in Table~\ref{tab:watermark_classification}. LLM watermarking techniques can be categorized into two types according to the point of watermark insertion: (1) Integrated into the model generation process~\citep{watermark_llm}, and (2) Post-processing after text generation~\citep{acl_muilit_bit, black_box_watermark}. 
Post-processing methods are independent of LLMs and they usually utilize masked language models (MLMs) (e.g. BERT~\citep{bert}, RoBERTa~\citep{roberta}) to replace tokens with synonyms. Both one-bit~\cite{black_box_watermark} or multi-bits~\cite{acl_muilit_bit} information can be injected in this manner.
% ~\cite{black_box_watermark} proposes injecting a watermark containing 1-bit information into the text generated by LLMs by performing synonym replacement.
% ~\cite{acl_muilit_bit} proposes a token substitution-based text watermarking method that achieves to encode multi-bits information into text from any source.
However, we argue that post-processing methods are not as effective as integrating methods for generating high-quality watermarked texts, since they do not take advantage of the generative power of LLMs. Moreover, replacement-based methods can only operate individual tokens and cannot adaptively change the subsequent generation sequence like integrating methods.
% which severely limits the freedom and consistency of the watermarked texts.

The most representative integrating
% ~\footnote{In this work, the LLM watermarking refers to the integrating method unless otherwise specified.} 
watermarking technique for LLMs is proposed by~\citet{watermark_llm}, which watermarks the LLM-generated text by manipulating the available part of the vocabulary. 
% Specifically, when LLM generates each token, it first creates a random seed based on the hash value of the previous token and splits the whole vocabulary into two parts (i.e., green list and red list) conditioned on that random seed. 
% Then, the logits of tokens in the green list will be added with a small positive value, which leads to that these tokens will be more likely to be sampled. In the watermark detecting stage, the same hash function is utilized to determine the number of tokens in the red and green list in a text, thereby determining whether the text contains watermark information. 
Following~\citet{watermark_llm}, several studies explore effective text watermarking in various scenarios.
~\citet{reliability_of_watermark} later improves this method by exploring diverse choices of random seed generators for splitting the vocabulary. 
~\citet{sweet} takes a step further to actively select high entropy tokens to watermark instead of all tokens, which is shown to be more effective in the code detection task. 
% as it will not affect the generation of those deterministic code segments.
~\citet{provable_robust_watermark} simplify the watermark method of~\citet{watermark_llm} by making the vocabulary splitting independent of the previously generated tokens and only dependent on a global key. 
However, all these methods generate watermarks that only contain 1 bit of information, which cannot meet the need to inject watermarks with diverse customized information.
There are some very concurrent works~\citep{advcaning_beyond, three_bricks} that focus on this problem. These methods still follow the random vocabulary partition~\citep{watermark_llm}, posing an unignorable threat to the quality of generated text caused by the multi-bits watermark.
In contrast, our Balance-Marking method can effectively maintain the text quality after injecting complex messages by balancing the probability distribution of the vocabulary partitions.

% In this work, we first analyze the research objectives of codable text watermarking, and then propose a practical method that achieves a good balance among multiple indicators designed for evaluating the effectiveness of CTWL.

\section{Mathematical Formulation of Codable LLM Watermarking}
\label{sec: format}
\subsection{Notations and Preliminaries}

Here, we first introduce the necessary notations used in this paper. Assume there is a large language model $LLM$ that takes a prompt sequence $\mathbf{x}^{prompt}$ as the input and sequentially outputs the corresponding tokens to form natural sentences as the response. At $l$-th step ($l=1,2,\cdots,L$), the entire input for $LLM$ is the combination of the original prompt $\mathbf{x}^{prompt}$ and the sequence of tokens $\mathbf{t}_{:(l-1)}=\{ t_{0},\cdots,t_{l-1}\}$ that is already predicted by $LLM$ in previous steps.\footnote{In the first step, there is no previously generated token, thus define $t_{0}$ is none.} 
Then, $LLM$ produces a probability distribution over the entire vocabulary $\mathcal{V}$ as $\mathbf{P}_{LLM}(\mathbf{x}^{prompt},\mathbf{t}_{:(l-1)})=(\cdots,P_{LLM}(v | \mathbf{x}^{prompt},\mathbf{t}_{:(l-1)}),\cdots )$, in which $P_{LLM}(v | \mathbf{x}^{prompt},\mathbf{t}_{:(l-1)})$ is the predicted probability of a specific token $v$ by $LLM$. The next token $t_{l}$ is sampled based on $\mathbf{P}_{LLM}(\mathbf{x}^{prompt},\mathbf{t}_{:(l-1)})$ according to specific sampling rules, such as multinomial sampling or greedy sampling.

\subsection{Formulation of Codable Text Watermarking}

\label{sec:formulation_of_codable_text_watermarking}

In this paper, we propose the concept of \textbf{codable text watermarking for LLMs} (\textbf{CTWL}) that can encode rich and necessary information into the text watermarks, in order to satisfy the multiple demands on the realistic application of LLMs. Formally, CTWL can be formulated as a \textit{message encoding} stage with a \textit{message decoding/extracting} stage:
\begin{equation}
\label{eq: codable watermark}
\begin{aligned}
 \textit{Encoding: } \qquad &  \mathcal{P}  \times \mathcal{M}  \rightarrow \mathcal{T}, \quad Enc(\mathbf{x}^{prompt},m) = \mathbf{t}, \\
 \textit{Decoding: } \qquad & \mathcal{T} \rightarrow \mathcal{M}, \quad Dec(\mathbf{t}) = m,
\end{aligned}
\end{equation}
where $\mathcal{P}$, $\mathcal{T}$ and $\mathcal{M}$ represents the prompt space, text space, and message space separately, $m \in \mathcal{M}$ is the message that needs to be encoded. Thus, one-bit watermarking methods can be considered as a simplified case of CTWL, where the message space only contains  \{0, 1\}.
Following the definition in Eq.~(\ref{eq: codable watermark}), the target on decoding messages from $\mathbf{t}$ can be written as 
\begin{equation}
\label{eq: target on extracting messages}
\begin{aligned}
 m = \mathop{\arg\max}_{m' \in \mathcal{M}}  P_{w}(m'|\mathbf{t})  ,
\end{aligned}
\end{equation}
where we need to design a specific probability function $P_{w}$ referred to as \textbf{message function} (refer to Section~\ref{sec: design of P}.) to measure how likely is that the watermarked message is $m'$ given $\mathbf{t}$.

Therefore, according to Bayes Formula, in the message encoding phase, it is equivalent to achieve\footnote{The detailed derivations can be found in Appendix~\ref{appendix: math derivations}.}
\begin{equation}
\label{eq: target on encoding messages}
\begin{aligned}
 & \mathop{\max}\limits_{\mathbf{t}}  \{ P_{w}(\mathbf{t}|m) / 
 \mathop{\max}_{m' \neq m} P_{w}(\mathbf{t}|m')  \}
 %\\ &\iff
 % \mathop{\max}\limits_{\mathbf{t}} \{ \mathop{\Pi}\limits_{l=1}^{L} P_{w}(t_{l}|,m,\mathbf{t}_{:(l-1)}) / \mathop{\max}_{m' \neq m} \mathop{\Pi}\limits_{l=1}^{L} P_{w}(t_{l}|m',\mathbf{t}_{:(l-1)})  \} 
 \\ &\iff
 \mathop{\max}\limits_{\mathbf{t}} \{ \mathop{\sum}\limits_{l=1}^{L} \mathop{\log}P_{w}(t_{l}|m,\mathbf{t}_{:(l-1)}) - \mathop{\max}_{m' \neq m} \mathop{\sum}\limits_{l=1}^{L} \mathop{\log}P_{w}(t_{l}|m',\mathbf{t}_{:(l-1)})  \} .
\end{aligned}
\end{equation}
%where $S_{enc}$ is the corresponding score function in the encoding phase to reflect the probability of the appearance of the text $\mathbf{t}$ given the message $m$.
That is, we aim to enlarge the gap between the probability that the text $\mathbf{t}$ is generated under message $m$ and the probability that it is generated under other $m'$.\footnote{We take the form of division in Eq.~(\ref{eq: target on encoding messages}) in order to successfully derive the optimization target to Eq.~(\ref{eq: encoding optimization problem}).} 
% Moreover, taking the quality of generated text as consideration, the encoding phase can be formulated as
% \begin{equation}
% \label{eq: encoding message and text quality}
% \begin{aligned}
% & \mathop{\max} \{ \mathop{\sum}\limits_{l=1}^{L} \mathop{\log}P(t_{l}|m,t_{:(l-1)}) - \mathop{\max}_{m' \neq m} \mathop{\sum}\limits_{l=1}^{L} \mathop{\log}P(t_{l}|m',t_{:(l-1)})  \} , 
%  \\
% & \text{s.t. \quad } \text{PPL}(t|\mathbf{x}^{prompt},m) \leq \text{PPL}(t| \mathbf{x}^{prompt}) +\epsilon ,
% \end{aligned}
% \end{equation}
% where we use perplexity (PPL) metric to measure the quality of the generated text, and aim to make the watermarked text have comparable quality with that of the text generated without embedded watermarks.
%Moreover, considering that the quality of the generated text with watermarks should not degrade greatly, we take the text generated without embedded watermarks $\mathbf{t}^{original}$ as a baseline for comparison. In this circumstance, we can express the encoding phase as:
However, as we can see, the above equation only considers the target of effectively hiding message $m$ into $\mathbf{t}$, but does not take the quality of the generated text into consideration. Thus, we take the original text generated without embedded watermarks $\mathbf{t}^{ori}$ as a baseline for comparison, and reformulate the encoding phase as:
\begin{equation}
\label{eq: encoding message and text quality original}
\begin{aligned}
& \mathop{\max}\limits_{\mathbf{t}} \{ \mathop{\sum}\limits_{l=1}^{L} \mathop{\log}P_{w}(t_{l}|m,\mathbf{t}_{:(l-1)}) - \mathop{\max}_{m' \neq m} \mathop{\sum}\limits_{l=1}^{L} \mathop{\log}P_{w}(t_{l}|m',\mathbf{t}_{:(l-1)})  \} , 
 \\
& \text{s.t. \quad } \text{PPL}(\mathbf{t}|\mathbf{x}^{prompt}) \leq \text{PPL}(\mathbf{t}^{ori}| \mathbf{x}^{prompt}) +\epsilon .
\end{aligned}
\end{equation}
Here, we utilize the perplexity (PPL) metric to measure the quality of the generated text, with the aim of ensuring the watermarked text maintains a similar quality to the text without watermarks. %As $\text{PPL}(\mathbf{t}|\mathbf{x}^{prompt})=[\mathop{\Pi}\nolimits_{l=1}^{L}P(t_{l}|\mathbf{x}^{prompt},\mathbf{t}_{:(l-1)})]^{-\frac{1}{L}} $, Eq.~(\ref{eq: encoding message and text quality original}) is equivalent to 
% \begin{equation}
% \label{eq: encoding message and text quality}
% \begin{aligned}
% & \mathop{\max}\limits_{\mathbf{t}} \{ \mathop{\sum}\limits_{l=1}^{L} \mathop{\log}P_{w}(t_{l}|m,\mathbf{t}_{:(l-1)}) - \mathop{\max}_{m' \neq m} \mathop{\sum}\limits_{l=1}^{L} \mathop{\log}P_{w}(t_{l}|m',\mathbf{t}_{:(l-1)})  \} , 
%  \\
% & \text{s.t. \quad } -\frac{1}{L}[\mathop{\sum}\limits_{l=1}^{L} \log P(t_{l}|\mathbf{x}^{prompt},\mathbf{t}_{:(l-1)})] \leq %-\frac{1}{L^{ori}}[\mathop{\Pi}\limits_{l=1}^{L^{ori}}P(t_{l}^{ori}|\mathbf{x}^{prompt},\mathbf{t}_{:(l-1)}^{ori})] +\epsilon .
% \log (\text{PPL}(\mathbf{t}^{ori}| \mathbf{x}^{prompt})  +\epsilon )
% \end{aligned}
% \end{equation}

\section{Balance-Marking: A Simple yet Effective CTWL Method}
\label{sec: method}

In this section, we first present an approximation algorithm to solve Eq.~(\ref{eq: encoding message and text quality original}) in Section~\ref{sec: encoding framwwork}, which serves as a general encoding algorithm for any predefined $P_{w}$. Then, we introduce two designs of $P_{w}$ in Section~\ref{sec: design of P} as unified guidelines to encode and decode messages. Finally, given the pre-defined encoding algorithm and $P_{w}$, the message extracting can be performed as presented in Section~\ref{subsec: message extracting phase}.

\subsection{A general framework for codable watermark encoding}
\label{sec: encoding framwwork}

In order to solve the constrained optimization problem in Eq.~(\ref{eq: encoding message and text quality original}), we are motivated to apply the method of Lagrange Multipliers by introducing a dual variable $\lambda$ and turn it into an unconstrained optimization problem. Furthermore, assume that the PPL scores in Eq.~(\ref{eq: encoding message and text quality original}) are calculated based on the same $LLM$ used to generate $t$,\footnote{This assumption is practical as the LLM is powerful enough to accurately measure the perplexity of a text.} given the prompt $\mathbf{x}^{prompt}$ and $LLM$, the term $ \text{PPL}_{LLM}(\mathbf{t}^{ori}| \mathbf{x}^{prompt})$ can be regarded as a constant. Therefore, 
%by using $\text{PPL}(\mathbf{t}|\mathbf{x}^{prompt})=[\mathop{\Pi}\nolimits_{l=1}^{L}P_{LLM}(t_{l}|\mathbf{x}^{prompt},\mathbf{t}_{:(l-1)})]^{-\frac{1}{L}} $, 
the target can be rephrased as:\footnotemark[3]
\begin{equation}
\label{eq: encoding optimization problem}
\begin{aligned}
\mathop{\max}_{\mathbf{t}} & \mathop{\sum}\limits_{l=1}^{L}\{\log P_{LLM}(t_{l}|\mathbf{x}^{prompt},\mathbf{t}_{:(l-1)})+\delta(\mathop{\log}P_{w}(t_{l}|m,\mathbf{t}_{:(l-1)}) - \mathop{\log}P_{w}(t_{l}|\hat{m},\mathbf{t}_{:(l-1)})\},
% & \text{s.t. \quad } \text{PPL}(t|\mathbf{x}^{prompt},m) \leq \min\limits_x\text{PPL}(x| \mathbf{x}^{prompt}) +\epsilon ,
\end{aligned}
\end{equation}
where we use $\text{PPL}(\mathbf{t}|\mathbf{x}^{prompt})=[\mathop{\Pi}\nolimits_{l=1}^{L}P_{LLM}(t_{l}|\mathbf{x}^{prompt},\mathbf{t}_{:(l-1)})]^{-\frac{1}{L}} $, and set $\delta=\frac{L}{\lambda}$ and $\hat m=\mathop{\mathrm{argmax}}_{m' \neq m} \mathop{\sum}\limits_{l=1}^{L} \mathop{\log}P(t_{l}|m',t_{:(l-1)})$.
% Inspired by the fact that the text decoding algorithms like beam search can use $\mathop{\log} P_{LLM}(x_{l}|\mathbf{x}^{prompt},x_{:(l-1)})$ to get a approximation solution of 
%Note that if without the watermarking requirement, existing text generation algorithms such as greedy search exactly aim to approximate the solution of $\mathop{\max}_{\mathbf{t}}\mathop{\sum}\nolimits_{l=1}^{L}\mathop{\log}P_{LLM}(t_{l}|\mathbf{x}^{prompt},\mathbf{t}_{:(l-1)}) $ by sampling the token based on the model logits $P_{LLM}(t_{l}|\mathbf{x}^{prompt},\mathbf{t}_{:(l-1)})$ during each step.
%if we substitute the term $\mathop{\log}P_{LLM}(x_{l}|\mathbf{x}^{prompt},x_{:(l-1)}) \}$ with $\mathop{\log}P_{LLM}(t_{l}|\mathbf{x}^{prompt},t_{:(l-1)}) + \delta(\mathop{\log}P(t_{l}|m,t_{:(l-1)}) - \mathop{\log}P(t_{l}|\hat{m},t_{:(l-1)})$, we can leverage the original text generation algorithm to tackle Eq.~\ref{eq: encoding message and text quality method}, thus accomplishing the encoding process. Furthermore, as solving $\hat m$ is challenging, for approximation, we loosen $\mathop{\log}P(t_{l}|\hat{m},t_{:(l-1)})$ to an approximation using $\frac{1}{|\mathcal{M}|}\mathop{\sum}\limits_{m'\in \mathcal{M}}\mathop{\log}P(t_{l}|m',t_{:(l-1)})$.
%Therefore, 

Eq.~(\ref{eq: encoding optimization problem}) motivates us that in order to encode $m$ into $\mathbf{t}$, we can manipulate the output logits during each token's generation by adding a term $\delta(\mathop{\log}P_{w}(t_{l}|m,\mathbf{t}_{:(l-1)}) - \mathop{\log}P_{w}(t_{l}|\hat{m},\mathbf{t}_{:(l-1)}))$ to the original log logits. \textbf{However, in practice, solving $\hat m$ is infeasible because the true $\hat{m}$ can only be solved after the whole output $t$ is determined, while we need to calculate $\delta(\mathop{\log}P_{w}(t_{l}|m,\mathbf{t}_{:(l-1)}) - \mathop{\log}P_{w}(t_{l}|\hat{m},\mathbf{t}_{:(l-1)})$ in each generation step according to Eq.~(\ref{eq: encoding optimization problem}).} Therefore, we replace $\mathop{\log}P_{w}(t_{l}|\hat{m},\mathbf{t}_{:(l-1)})$ with calculable $\frac{1}{|\mathcal{M}|}\mathop{\sum}\nolimits_{m'\in \mathcal{M}}\mathop{\log}P_{w}(t_{l}|m',\mathbf{t}_{:(l-1)})$ as an alternative, and finally get the message encoding object function in each generation step as:
\begin{equation}
\begin{aligned}
    L(m,\mathbf x^{prompt}&, 
    \mathbf{t}_{:(l-1)})= \mathop{\max}\limits_{v}\{\underbrace{\mathop{\log}P_{LLM}(v|\mathbf{x}^{prompt},\mathbf{t}_{:(l-1)})}_{\text{model logit}}\\ &+ \delta(\underbrace{\mathop{\log}P_{w}(v|m,\mathbf{t}_{:(l-1)}) - \frac{1}{|\mathcal{M}|}\mathop{\sum}\limits_{m'\in \mathcal{M}}\mathop{\log}P_{w}(v|m',\mathbf{t}_{:(l-1)}}_{\text{message logit}})\},
    \label{eq: encoding objective function}
\end{aligned}
\end{equation}
where we denote the first term in the right as the \textbf{model logit}, which is determined by the $LLM$ only; we denote the second additional term as the \textbf{message logit}, which is the key component for encoding message $m$ into $t$.

As we can see, as long as the function $P_{w}$ is well-defined, the encoding process can be completed by adding the message logits to the model logits and sampling the token based on the new logits. We put the general message encoding procedure in Algorithm~\ref{algo: encoding algorithm for general P},\footnote{Though we employ a greedy search as the text generation algorithm in Algorithm~\ref{algo: encoding algorithm for general P} for example, our framework is also compatible with other generation rules such as beam search.} and will discuss how to properly design $P_{w}$ in detail in the next section.

% In summary, the algorithm for encoding can be realized by adjusting the model logits used in text generation procedures such as beam search, i.e. adding to logits an additional term $\delta(\mathop{\log}P(t_{l}|m,t_{:(l-1)}) - \frac{1}{|\mathcal{M}|}\mathop{\sum}\limits_{m'\in \mathcal{M}}\mathop{\log}P(t_{l}|\mathbf{x}^{prompt},m',t_{:(l-1)})$. 
% Algorithm~\ref{algo: encoding algorithm for general P} describes the detailed steps of the encoding algorithm for a settled $P_{w}$. For simplicity, we employ a greedy search as the text generation algorithm in our demonstration.
% In summary, we can implement the encoding algorithm by manipulating the model logits used in text generation algorithms, such as beam search. This manipulation involves the addition of a term, $\delta(\mathop{\log}P(t_{l}|m,t_{:(l-1)}) - \frac{1}{|\mathcal{M}|}\mathop{\sum}\limits_{m'\in \mathcal{M}}\mathop{\log}P(t_{l}|m',t_{:(l-1)})$, to the original logits.
% Algorithm~\ref{algo: encoding algorithm for general P} describes the detailed steps of the encoding algorithm for a settled $P_{w}$. For simplicity, we employ a greedy search as the text generation algorithm in Algorithm~\ref{algo: encoding algorithm for general P}.

\begin{algorithm}[t]
\SetAlgoLined
% \KwResult{Result of the algorithm}
\KwIn{Language model $LLM$, prompt $\mathbf x^{prompt}$, message $m$, watermarking weight $\delta$}
% \KwOut{Here, provide the expected output/results of your algorithm}
%$\mathbf{x}^{:-1}:=\mathbf{x}^{prompt}$ \;
\For{$l=1,\cdots, L$}{

1. Calculate $\mathop{\log}P_{LLM}(v|\mathbf{x}^{prompt},\mathbf{t}_{:(l-1)}) \}$ for each $v$ in the vocabulary using $LLM$\;
2. Calculate $\mathop{\log}P_{w}(v|m,\mathbf{t}_{:(l-1)})$ based on the settled $P_{w}$\; 
3. Select $t_{l} = \mathop{\arg\max}\limits_{v}\ \{\mathop{\log}P_{LLM}(v|\mathbf{x}^{prompt},\mathbf{t}_{:(l-1)}) + \delta(\mathop{\log}P_{w}(v|m,\mathbf{t}_{:(l-1)}) - \frac{1}{|\mathcal{M}|}\mathop{\sum}\limits_{m'\in \mathcal{M}}\mathop{\log}P_{w}(v|m',\mathbf{t}_{:(l-1)})\}$
}
\KwOut{watermarked text $\mathbf{t} = \{t_1,t_2,\cdots,t_{L}\}$}
\caption{A General Message Encoding Framework for A Settled $P_{w}$}
\label{algo: encoding algorithm for general P}
\end{algorithm}

\subsection{The Design of message function $P_{w}$}
\label{sec: design of P}

% In Section~\ref{sec: encoding framwwork}, we propose a general encoding algorithm for $P_{w}$. Also, the decoding can be directly performed given $P_{w}$. So, the only thing that needs discussion is the design of $P_{w}$. Considering Algorithm~\ref{algo: encoding algorithm for general P}, a good design of P is supposed to make the following term $L(m,\mathbf x^{prompt}, \mathbf{t}_{:(l-1)})$ as large as possible for any message $m$:
In the above section, we present a general encoding algorithm for an arbitrary $P_{w}$. %Consequently, the focus of our discussion shifts to the design of $P_{w}$. 
In the following, we will introduce two designs of $P_{w}$ as our preliminary attempts toward CTWL. 

\subsubsection{Vanilla $P_{w}$ for Random Vocabulary Partition}
\label{sec: vanilla P}
According to Eq.~(\ref{eq: encoding objective function}), a high value of $L(m,\mathbf x^{prompt}, \mathbf{t}_{:(l-1)})$ relies on the existence of a $v$ with a high message logit. In other words, there should be a $v$ for which $\log P_{w}(v|m,\mathbf{t}_{:(l-1)})$ greatly surpasses the mean. To achieve this, one natural idea is to ensure that the distribution $\mathbf P_{w}(m,\mathbf{t}_{:(l-1)}) = (P_{w}(v_1|m,\mathbf{t}_{:(l-1)}),P_{w}(v_2|m,\mathbf{t}_{:(l-1)}),\cdots,P_{w}(v_{|\mathcal{V}|}|m,\mathbf{t}))$ varies greatly across distinct messages. In this way, for a message $m$, there would at least exist a $v$ whose $\log P_w(v|m,\mathbf{t}_{:(l-1)})$ deviates far from the mean $\frac{1}{|\mathcal{M}|}\mathop{\sum}\nolimits_{m'\in \mathcal{M}}\mathop{\log}P_{w}(v|m',\mathbf{t}_{:(l-1)})$, thereby resulting in a high message logit. To ensure such differences in $\mathbf P_{w}(m,\mathbf{t}_{:(l-1)})$ with different $m$, we can assign random values for $\mathbf P_{w}(m,\mathbf{t}_{:(l-1)})$ based on the random seeds directly decided by their own $m$:
% One straightforward method to design $P_{w}$ is to assign random values for $\mathop{\log}P(v|m,x_{:(l-1)})$. This random assignment allows for the existence of a certain $v$ that results in a large $\mathop{\log}P(v|m,x_{:(l-1)}) - \frac{1}{|\mathcal{M}|}\mathop{\sum}\limits_{m'\in \mathcal{M}}\mathop{\log}P(v|m',x_{:(l-1)})$ for any given message. This vanilla approach can be seen as an expansion of the soft watermarking method proposed by~\citet{watermark_llm}, which we refer to as \textbf{Vanilla Codable Text Watermarking for LLMs~(Vanilla CTWL)}\wang{how to name it?}.
\begin{equation}
    \label{eq: vanilla_CTWL raw}
  \log \hat{P}_w(v|m,\mathbf{t}_{:(l-1)}) = \begin{cases}
        1 , & h(v,m,\mathbf{t}_{:(l-1)}) = 1, \\
        0 ,& h(v,m,\mathbf{t}_{:(l-1)}) = 0.
    \end{cases}
\end{equation}
\begin{equation}
    \label{eq: vanilla_CTWL}
  \log P_w(v|m,\mathbf{t}_{:(l-1)}) = \log \frac{\hat{P}_w(v|m,\mathbf{t}_{:(l-1)})}{\sum\limits_{v} \hat{P}_w(v|m,\mathbf{t}_{:(l-1)})}.
\end{equation}
In the above, $h$ denotes a hash function that maps the input $(v,m,\mathbf{t}_{:(l-1)})$ to either 0 or 1. 
%, and $Z$ is a normalization term to ensure that $\mathop{\sum}\limits_{v} P_w(v|m,x_{:(l-1)}) = 1$. 
This can be considered as a vanilla extension from the soft watermarking method in~\citet{watermark_llm} by further taking $m$ into consideration, thus we denote it as \textbf{Vanilla-Marking}.

%Additionally, because the assignment of values in $P_w$ is random, $\frac{1}{|\mathcal{M}|}\mathop{\sum}\nolimits_{m'\in \mathcal{M}}\mathop{\log}P_w(v|m',\mathbf{t}_{:(l-1)})$ is nearly constant, so we can omit it in the encoding procedure.

\subsubsection{$LM_{proxy}$-aided $P_{w}$ for Balance Vocabulary Partition}

\label{subsubsec: balance-marking}
% The vanilla $P_{w}$ in Section~\ref{sec: vanilla P} does not incorporate $\mathop{\log}P_{LLM}(v|\mathbf{x}^{prompt},\mathbf{t}_{:(l-1)})$, potentially resulting in a suboptimal balance between watermark and text quality. For any given message $m$, to ensure that the whole $L(m,\mathbf x^{prompt}, \mathbf{t}_{:(l-1)})$ value is large, there needs to be a particular $v$ satisfying both high $\mathop{\log}P_{LLM}(v|\mathbf{x}^{prompt},\mathbf{t}_{:(l-1)})$ and high $\mathop{\log}P(v|m,x_{:(l-1)}) - \frac{1}{|\mathcal{M}|}\mathop{\sum}\limits_{m'\in \mathcal{M}}\mathop{\log}P(v|m',x_{:(l-1)})$. Encouraged by this insight, for each message $m$, we randomly choose a set $V_{m,\mathbf{t}_{:(l-1)}}$ from the vocabulary that satisfies the below condition:

%The message function $P_w$ proposed in Section~\ref{sec: vanilla P} indeed ensures the existence of a $v$ with a high message logit. However, 
%the same $v$ may have a low model logit, which will lead to a low $L(m,\mathbf x^{prompt}, \mathbf{t}_{:(l-1)})$. 
The message function $P_w$ proposed in Section~\ref{sec: vanilla P} has a problem that it does not guarantee that the same $v$ can also have a high model logit at the same time. 
This could potentially result in a small sum of the model logit and the message logit. 
Therefore, we argue that \textbf{a more advanced $P_{w}$ should produce a $v$ with both a high model logit and a message logit}. 

To accomplish this, we are motivated to utilize the model logit distribution $\mathbf{P}_{LLM}(\mathbf{x}^{prompt},\mathbf{t}_{:(l-1)})$ as prior knowledge, and pre-select a subset of tokens that is likely to contain some tokens with high model logits in advance. Then, we assign high message logits to the tokens in the above subset, ensuring the existence of a token $v$ with both a high model logit and message logit. Formally, we propose Algorithm~\ref{algo: choose V_{m,x_{:(l-1)}}_old} to randomly choose the subset $V_{m,\mathbf{t}_{:(l-1)}}$ that satisfies the following condition:
% \begin{equation}
% \label{eq: V_m condition}
%     \mathop{\sum}\limits_{v\in V_{m,\mathbf{t}_{:(l-1)}}} P_{LM_{proxy}}(v| x_{(l-1- L_{prefix}):(l-1)}) \geq 0.5 
% \end{equation}
\begin{equation}
\label{eq: V_m condition}
    \mathop{\sum}\limits_{v\in V_{m, \mathbf{t}_{:(l-1)}}}P_{LLM}(v|\mathbf{x}^{prompt},\mathbf{t}_{:(l-1)}) \geq \sigma, 
\end{equation}
where $\sigma$ is a controllable threshold. W set $\sigma=0.5$ in the following paper unless otherwise stated, because we believe that balancing the probability accumulations of tokens within and out of $V_{m, \mathbf{t}_{:(l-1)}}$ can achieve the maximal diversity of $V_{m, \mathbf{t}_{:(l-1)}}$ w.r.t.\ different $m$. The reason why we design Algorithm~\ref{algo: choose V_{m,x_{:(l-1)}}_old} in such a way is, \textbf{the case when all $\{P_{LLM}(v|\mathbf x^{prompt}, \mathbf{t}_{:(l-1)})|v \in V_{m,\mathbf{t}_{:(l-1)}}\}$ values tend to be small, yet still sum to $\sigma$, is very unlikely to occur}. Thus, there should always be some $P_{LLM}(v|\mathbf x^{prompt}, \mathbf{t}_{:(l-1)})$ that is relatively large to make the summation exceed the threshold. Additionally, introducing randomness in the selection process of $V_{m,\mathbf{t}_{:(l-1)}}$ can enlarge the difference in $V_{m,\mathbf{t}_{:(l-1)}}$ among different messages $m$, which plays the same role as that in Vanilla-Marking. 

After getting $V_{m, \mathbf{t}_{:(l-1)}}$, we assign the message logits by following Eq.(~\ref{eq: vanilla_CTWL}) but modifying Eq.~(\ref{eq: vanilla_CTWL raw}) as
\begin{equation}
    \label{eq: LM_pub-aided P raw}
     \log \hat{P}_w(v|m,\mathbf{t}_{:(l-1)}) = \begin{cases}
        1 , & v \in V_{m,\mathbf{t}_{:(l-1)}}, \\
        0 ,& v \not\in V_{m,\mathbf{t}_{:(l-1)}}.
    \end{cases}
\end{equation}

\begin{algorithm}[t]
\SetAlgoLined
% \KwResult{Result of the algorithm}
\KwIn{Message $m$, text prefix $\mathbf{t}_{:(l-1)}$, language model $LLM$, \textcolor{red}{proxy-LM} $\textcolor{red}{LM_{proxy}}$, $\textcolor{red}{\mathcal{M_A}=\{1,\cdots,A\}}$, hash functions $h$ and $\textcolor{red}{\hat{h}}$.}
% \KwOut{Here, provide the expected output/results of your algorithm}
1. Calculate a seed $s = h(m,\mathbf{t}_{:(l-1)})$, or $\textcolor{red}{s = h(\hat h(m),\mathbf{t}_{:(l-1)})}$ \textcolor{red}{where} $\textcolor{red}{\hat h}$ \textcolor{red}{maps} $\textcolor{red}{m}$ \textcolor{red}{to} $\textcolor{red}{\hat{h}(m) \in \mathcal{M}_{A}}$\;
2. Shuffle the vocab list $(v_1, \cdots, v_{|\mathcal{V}|})$ to $(v'_1, \cdots, v'_{|\mathcal{V}|})$ with the seed $s$\;
3. Select the first $k$ tokens in the shuffled list so that $k$ is the minimal value to make $\{v'_1,\cdots,v'_k\}$ satisfy Eq.~(\ref{eq: V_m condition}) or \textcolor{red}{Eq.~(\ref{eq: V_m condition proxy})}.

\KwOut{$V_{m,\mathbf{t}_{:(l-1)}} = \{v'_1,\cdots,v'_k\}$}
\caption{Algorithm of Choosing Subset $V_{m,\mathbf{t}_{:(l-1)}}$ \textcolor{red}{(Practical Version in Red)}}
\label{algo: choose V_{m,x_{:(l-1)}}_old}
\end{algorithm}

% However, such a method faces two challenges. (1) Since $P_{w}$ is a function of the input $(m, x_{:(l-1)})$, it cannot involve $\mathbf{x}^{prompt}$ (2) During the decoding process, we may only receive a segment of the watermarked text, leading to disparities between the $x_{:(l-1)}$ used in encoding and decoding. Given this, we use $P_{LLM}(v| \mathbf{t}_{(l-1- L_{prefix}):(l-1)})$ to estimate $P_{LLM}(v|\mathbf{x}^{prompt},\mathbf{t}_{:(l-1)})$, where we omit $\mathbf{x}^{prompt}$, and truncate $x_{:(l-1)}$ to a fixed-length $x_{(l-1- L_{prefix}):(l-1)}$, so that it can remain consistent during encoding and decoding.
%Nevertheless, the above approach still encounters several obstacles in realistic applications. Thus, 
Also, we make some specific improvements to make our method more practical in various scenarios. We summarize these strategies briefly here, and put the detailed illustrations in Appendix~\ref{appendix: improvements of balance-marking}: \textbf{(1)} We omit $\mathbf{x}^{prompt}$ and truncate $\mathbf{t}_{:(l-1)}$ to a fixed-length $\mathbf{t}_{(l-1- L_{prefix}):(l-1)}$ for consistency during both encoding and decoding. 
\textbf{(2)} To make our method applicable in various scenarios discussed in Appendix~\ref{sec: scenes}, we broaden the $LLM$ used in $P_{LLM}(v| \mathbf{t}_{(l-1- L_{prefix}):(l-1)})$ into a general proxy model denoted as $LM_{proxy}$, and modify the condition in Eq.~(\ref{eq: V_m condition}) to:
\begin{equation}
\label{eq: V_m condition proxy}
    \mathop{\sum}\limits_{v\in V_{m, \mathbf{t}_{:(l-1)}}} P_{LM_{proxy}}(v| \mathbf{t}_{(l-1- L_{prefix}):(l-1)}) \geq \sigma.
\end{equation}
%\paragraph{(3) Pre-map message space into a smaller space for efficient computing.} 
%Changing the condition Eq.~(\ref{eq: V_m condition}) in Algorithm~\ref{algo: choose V_{m,x_{:(l-1)}}_old} to the condition Eq.~(\ref{eq: V_m condition proxy}) results in Algorithm~\ref{algo: choose V_{m,x_{:(l-1)}}}. 
%Since computing $V_{m,\mathbf{t}_{:(l-1)}}$ for each $m$ during encoding can be much time-consuming when the message space is pretty large, 
%Considering the entire message space may be pretty large and for efficient encoding and decoding, 
%we opt to first map the entire message space into a smaller space as $m \rightarrow \hat{h}(m) \in \mathcal{M}_{A}=\{1,\cdots, A \}$\footnote{For a detailed discussion about the choice of $A$, refer to Section~\ref{sec: The effect of $A$}.} by using another hash function $\hat{h}$, and then compute the seed $s$ as $s = h(\hat h(m), x_{:(l-1)})$. 
%By this way, we only need to run Algorithm~\ref{algo: choose V_{m,x_{:(l-1)}}_old} a mere $A$ times for each $\mathbf{t}_{:(l-1)}$.
%\footnote{For a detailed discussion about the choice of $A$, refer to Section~\ref{sec: The effect of $A$}.}
% and use this condition in Algorithm~\ref{algo: choose V_{m,x_{:(l-1)}}_old} and Eq.~(\ref{eq: LM_pub-aided P}). 
\textbf{(3)} For efficient encoding and decoding on the entire message space, we opt to first map the entire message space into a smaller space as $m \rightarrow \hat{h}(m) \in \mathcal{M}_{A}=\{1,\cdots, A \}$\footnote{For a detailed discussion about the choice of $A$, refer to Appendix~\ref{sec: The effect of $A$}.} by using another hash function $\hat{h}$, and then compute the seed $s$ as $s = h(\hat h(m), x_{:(l-1)})$.

% \begin{algorithm}[t]
% \SetAlgoLined
% % \KwResult{Result of the algorithm}
% \KwIn{Message $m$, text prefix $\mathbf{t}_{:(l-1)}$, proxy-LM $LM_{proxy}$, $\mathcal{M_A}=\{1,\cdots,A\}$.}
% % \KwOut{Here, provide the expected output/results of your algorithm}
% 1. Calculate a seed $s = h(\hat h(m),\mathbf{t}_{:(l-1)})$ with a hash function $h$ and another hash function $\hat h$ that maps $m$ to $\hat{h}(m) \in \mathcal{M}_{A}$\;
% 2. Shuffle the vocab list $(v_1, \cdots, v_{|\mathcal{V}|})$ to $(v'_1, \cdots, v'_{|\mathcal{V}|})$ with the seed $s$\;
% 3. Select the first $k$ tokens in the shuffled list so that $k$ is the minimal value to make $\{v'_1,\cdots,v'_k\}$ satisfy Eq.~(\ref{eq: V_m condition proxy}).\

% \KwOut{$V_{m,\mathbf{t}_{:(l-1)}} = \{v'_1,...,v'_k\}$}
% \caption{Practical Version of Choosing Subset $V_{m,\mathbf{t}_{:(l-1)}}$}
% \label{algo: choose V_{m,x_{:(l-1)}}}
% \end{algorithm}

%The watermarking algorithm, which uses $P_{w}$ as defined in Eq.~(\ref{eq: LM_pub-aided P}) and Algorithm~\ref{algo: choose V_{m,x_{:(l-1)}}}, is referred to as \textbf{Balance-Marking}. 
%All in all, we summarize the final practical version of the above $LM_{proxy}$-aided watermarking method in Algorithm~\ref{algo: choose V_{m,x_{:(l-1)}}}. 
The practical version of the above $LM_{proxy}$-aided watermarking method is specially highlighted in red in Algorithm~\ref{algo: choose V_{m,x_{:(l-1)}}_old}. Given that it employs a probability-balanced vocabulary partition, we refer to it as \textbf{Balance-Marking}. 
%Still, we omit $\frac{1}{|\mathcal{M}|}\mathop{\sum}\limits_{m'\in \mathcal{M}}\mathop{\log}P_w(v|m',\mathbf{t}_{:(l-1)})$ during encoding as discussed in Section~\ref{sec: vanilla P}.
 % Additionally, thanks to the randomness in selecting, $\frac{1}{|\mathcal{M}|}\mathop{\sum}\limits_{m'\in \mathcal{M}}\mathop{\log}P_w(v|m',x_{:(l-1)})$ is still nearly a constant,  so we omit it in encoding as in Section~\ref{sec: vanilla P}.

%\subsubsection{An intuitive comparison of Vanilla-Marking and Balance-Marking}
%\label{sec: intuitive analysis}
%In the watermark algorithm proposed by \cite{watermark_llm}, the candidates for the preceding tokens are randomly split into two parts: the available token part (i.e., ``green token list'') and the unavailable token part (i.e., ``red token list''). The model is then encouraged to generate tokens lying in the green list so as to watermark the text. In our work, Eq.~(\ref{eq: vanilla_CTWL raw}) and Eq.~(\ref{eq: vanilla_CTWL}) of Vanilla-Marking, and Eq.~(\ref{eq: LM_pub-aided P raw}) and Eq.~(\ref{eq: LM_pub-aided P}) of Balance-Marking fulfill similar roles, assigning some tokens with higher message logits.

%Here, we make a comparison between Balance-Marking and Vanilla-Marking. First, \textbf{Balance-Marking is more effective in maintaining text quality 
%when choosing the available tokens for a specific message compared to Vanilla-Marking}.
Here, we briefly discuss about the two advantages that Balance-Marking has compared to Vanilla-Marking. (1) First, \textbf{Balance-Marking is more effective in maintaining text quality}. That is because the way of pre-choosing the subset $V_{m,\mathbf{t}_{:(l-1)}}$
ensures that Balance-Marking can find tokens with both high model logits and message logits, thus making the next generated word more meaningful and reliable. However, Vanilla-Marking, which just randomly selects the available tokens, might end up with an unreasonable next token.  (2) Secondly and interestingly, \textbf{Balance-Marking also has the ability to automatically bypass low-entropy sections of the text}, which is a property that \citet{sweet} explicitly aims to achieve. For example, if there is only one reasonable token candidate whose predicted probability by proxy-LM is almost 1.0, then for all messages, this token would be selected into the available part. In other words, this position in the sequence is implicitly ``skipped'' during watermark encoding and decoding, and the decoding process is actually carried out by comparing the values of $P_w$ under different $m$ on those high-entropy sections of the text.

%That is because Balance-Marking strives to find the available token list whose accumulated model logits surpass a certain percentage of the total probability sum. This strategy 
%That is because the way of pre-choosing the subset $V_{m,\mathbf{t}_{:(l-1)}}$ in Algorithm~\ref{algo: choose V_{m,x_{:(l-1)}}_old} 
%ensures that Balance-Marking can include some tokens with relatively high model logits in the available token list, and make the next word generation more reliable. However, Vanilla-Marking, which just randomly selects the available tokens, might end up with an unreasonable next token and reduce text quality. 
%This forces the model to generate either texts with high perplexity (PPL) or texts devoid of any watermark.

% Interestingly, \textbf{Balance-Marking also has the ability to automatically bypass low-entropy sections of the text}, which is a property that \citet{sweet} explicitly aims to achieve. For example, consider a situation in which there is only one reasonable token candidate, whose predicted probability by proxy-LM is almost 1.0. Then, 
%Take for example a situation where there is only one reasonable preceding token,  which possesses a probability nearing 1. 
% for all messages, this token would be selected into the available part, according to Eq.~(\ref{eq: V_m condition proxy}). In other words, this position in the sequence is implicitly ``skipped'' during watermark encoding and decoding, and the decoding process is actually carried out by comparing the values of $P_w$ under different $m$ on those high-entropy sections of the text.

\subsection{Message Decoding}
\label{subsec: message extracting phase}
For a settled $P_w$, the decoding process is conducted by finding a solution to Eq.~(\ref{eq: target on extracting messages}). As per Bayes' Theorem, this can be rewritten as:
\begin{equation}
\label{eq: decoding messages}
\begin{aligned}
 m &= \mathop{\arg\max}_{m' \in \mathcal{M}}  P_{w}(\mathbf{t}|m'), = \mathop{\arg\max}_{m' \in \mathcal{M}} \{ \mathop{\sum}\limits_{l=1}^{L} \mathop{\log}P_{w}(t_{l}|m',\mathbf{t}_{:(l-1)})\}.
\end{aligned}
\end{equation}
Eq.~(\ref{eq: decoding messages}) can be computed directly using either Eq.~(\ref{eq: vanilla_CTWL raw}) and  Eq.~(\ref{eq: vanilla_CTWL}) for Vanilla-Marking, or Eq.~(\ref{eq: LM_pub-aided P raw}) and  Eq.~(\ref{eq: vanilla_CTWL}) for Balance-Marking. 
%Similar to Section~\ref{sec: vanilla P}, we omit $Z_{m,\mathbf{t}_{:(l-1)}}$ in  Eq.~(\ref{eq: vanilla_CTWL}) and Eq.~(\ref{eq: LM_pub-aided P}) to simplify the computation, since $Z_{m,\mathbf{t}_{:(l-1)}}$ values are similar across different messages $m$, due to the randomness in the assignment of values to $P_w$.
Furthermore, when we need to encode multiple messages into text, we can sequentially encode each message into one segment of the text (e.g., every 100 tokens of the text). Then, message decoding can be conducted independently within each segment.

% ToDo: V2里将更多的数据集引入进来，将单纯的Evaluation System升级成包含Eval和Datasets的Benchmark
% \section{Benchmark of Codable Text Watermarking for LLMs}

% \newpage
\section{Experiment}
\label{sec: experiment}
\subsection{Experimental Settings}

\label{subsec: experimental settings}
% not compressed yet 9.26
Following the experimental settings used by~\citet{watermark_llm}, we utilize OPT-1.3B~\citep{Zhang2022OPTOP} for the generation of texts in main experiments. We also conduct experiments on LLaMA-7/13B~\citep{llama} models. 
Our prompt inputs are derived from the news-like subset of the C4 dataset~\citep{2019t5}. In each experiment, we extract 500 prompt inputs and truncate them to a uniform length of 300 tokens. The language model is then requested to generate 200 tokens via a 4-way beam search. 
To mitigate repetition in the generated text, we implement a repetition penalty of 1.5.
% , which is applied akin to the ``alpha\_presence'' parameter in the OpenAI API.\footnote{\url{https://platform.openai.com/docs/api-reference/parameter-details}}
We evaluate the quality of the text by perplexity. For OPT-1.3B, we use OPT-2.3B to calculate perplexity. As for LLaMA-7/13B, we use LLaMA-33B to calculate perplexity.
The evaluation protocol for CTWL is introduced in Appendix~\ref{sec: evaluation}. Besides, a case study is provided in Appendix~\ref{sec: examples}.

As for the $LM_{proxy}$ in Balance-Marking, we opt to use GPT2~\citep{gpt2}, a well-known, publicly available, and comparatively smaller language model, which comprises 124M parameters. Additional hyper-parameters used in Balance-Marking are set to the following: $A=100$, $L_{prefix}=10$, $\sigma=0.5$ and $\mathcal{M}=\{0,1,...,2^{20}-1\}$. Here, a message $m\in \mathcal{M}$ corresponds to 20-bit information. The hash scheme is the same as \citet{watermark_llm}. A detailed discussion is presented in~\ref{appendix: hyperprameters}. Besides, we conduct some approximation for acceleration, which is detailed in Appendix~\ref{appendix: reasons for omitting}.

\begin{figure}[t]
  \centering
  % \begin{subfigure}{.95\textwidth}
    \includegraphics[width=\linewidth]{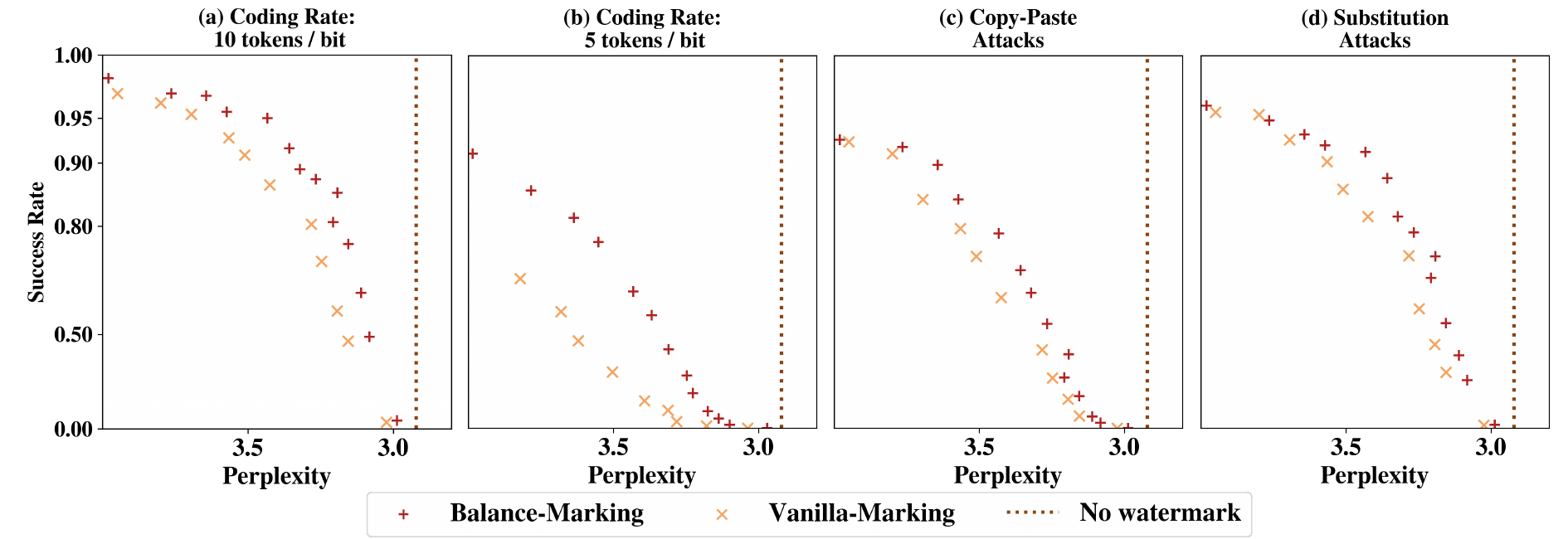}
    % \phantomsubcaption
    % \label{subfig:ppl-WSR-10}
  % \end{subfigure}%

 \begin{subfigure}{\textwidth} % this 'subfigure' env. has no visible content
            \refstepcounter{subfigure}\label{subfig:ppl-WSR-10}
            \refstepcounter{subfigure}\label{subfig:ppl-WSR-5}
              \refstepcounter{subfigure} \label{fig:cp}
                \refstepcounter{subfigure}\label{fig:sub-0.05}
\end{subfigure}%
  
%    \begin{subfigure}{0pt}
%     % \subcaption{Coding Rate: 5.}
%     % \subcaption{}
%      \phantomsubcaption
%     \label{subfig:ppl-WSR-5}
%   \end{subfigure}

%    \begin{subfigure}[b]{0pt}
%     \phantomsubcaption
%     \label{fig:cp}
%     \end{subfigure}
% %     \hfill
%     \begin{subfigure}[b]{0pt}  
%         \phantomsubcaption
%         \label{fig:sub-0.05}
%     \end{subfigure}
%     \begin{subfigure}[b]{0. \textwidth}  
%         \phantomsubcaption
%         \label{fig:legand}
%     \end{subfigure}
\caption{Balance-Marking outperforms Vanilla-marking under both different coding rates (subfigure a, b) and different attack scenarios (subfigure c, d).}
\end{figure}

% not compressed yet
\subsection{The Results of Watermark Quality}
\label{sec: main_results}

% For both Vanilla-Marking and Balance-Marking, we can adjust $\delta$ in Algorithm~\ref{algo: encoding algorithm for general P} to balance text quality and watermark success rate. A high $\delta$ encourages a strong watermark but hurts text quality. To explore the performance of a specific watermark algorithm, we run experiments with multiple $\delta$s (i.e., $\{0.5, 0.8, 0.9, 1.0, 1.1, 1.2, 1.3, 1.4, 1.5, 1.8, 2.0, 2.5, 3.0, 4.0\}$) and record the corresponding text quality and watermark success rate.

% Figure~\ref{subfig:ppl-WSR-10} and~\ref{subfig:ppl-WSR-5} illustrate the trade-off relationship between text quality (measured via perplexity) and the watermarking success rate of restoring the injected watermark message, Success$_\text{m}$, under different payload information coding rates. We filtered out the points with too poor text quality (PPL $> 4$) and rescaled the y-axis for clearer demonstration. As expected, an increase in Success$_\text{m}$ compromises text quality. Besides, watermarks with high coding rates, which encapsulate more information, are particularly challenging to embed into text without causing noticeable effects. Such watermarks tend to increase perplexity, degrading text quality. 

% Upon analysis, it is evident that Balance-Marking surpasses the Vanilla-Marking approach, achieving a more favorable balance between text quality and the success rate of watermarking.
We experiment with Vanilla-Marking and Balance-Marking by adjusting the parameter $\delta$ in Algorithm~\ref{algo: encoding algorithm for general P}. This parameter helps strike a balance between text quality and watermark success rate. Higher $\delta$ values lead to a stronger watermark but at the expense of text quality. We run tests using a range of $\delta$ values (i.e., $\{0.5, 0.8, 0.9, 1.0, 1.1, 1.2, 1.3, 1.4, 1.5, 1.8, 2.0, 2.5, 3.0, 4.0\}$) and coding rates of 10 or 5 tokens per bit to observe the corresponding effects. 

Figure~\ref{subfig:ppl-WSR-10} and~\ref{subfig:ppl-WSR-5} depict the trade-off between text quality, assessed by perplexity, and the success rate of recovering the embedded watermark. Enhancing the watermark's success rate tends to adversely affect text quality, and embedding watermarks with a high coding rate of 5 tokens per bit is particularly challenging. In both cases, \textbf{Balance-Marking outperforms Vanilla-Marking, offering a superior trade-off between text quality and watermark success rate.}

\subsection{The Results of Robustness to Real-world Attacks}
\label{subsec: robustness_attack}
% not compressed yet
% In practical use cases, the watermark embedded within a text may face attenuation due to various attacks. It may become challenging to detect when concealed within human-generated texts, a scenario termed as Copy-Paste Attacks~\citep{reliability_of_watermark}. Additionally, the watermark might be exposed to erosion stemming from activities like word substitution, which we refer to as Substitution Attacks.%\footnote{Both Vanilla-Marking and Balance-Marking are unable to decode the original message after Paraphrasing Attacks~\citep{reliability_of_watermark} performed by GPT-3.5-turbo. We discuss this as a limitation in Section~\ref{sec: limitation}.}. not finished. remove this due to removal of limitation section
% Here, we only apply these attacks to the watermark under the coding rate of 10 tokens per bit. This is based on our observation that a payload of 5 tokens per bit results in a watermark excessively susceptible to attacks.

In real-world applications, the embedded watermark in a text can be weakened by different types of attacks. For example, watermarks can be difficult to detect when hidden in human-written texts, a situation known as Copy-Paste Attacks~\citep{reliability_of_watermark}. Moreover, the watermark can be eroded by actions like word substitution, referred to as Substitution Attacks. In our study, we apply these attacks to the watermark under a coding rate of 10 tokens per bit.\footnote{We omit the case of coding rate of 5 tokens per bit since it makes both watermarks ineffective under attacks.}

% V2 ToDO: 可以考虑在这里画一张直观的图，展示我们的方法对于人类/机器文本边界的判断情况
% 直观的把窗口在混合文本上滑动时的结果置信度画出来
\paragraph{Robustness to Copy-paste Attacks.}
\label{sec: copy-paste attack}

A possible way to use LLM-generated text is to integrate it into human-written documents. We refer to this as a ``Copy-Paste Attack"~\citep{reliability_of_watermark}. This presents a challenge in watermark detection as the location of the watermarked text becomes unpredictable. To simulate this, we insert a 200-token watermarked text into a 1000-token piece of human-written text, taken from the C4 dataset. The detection process involves using a sliding window technique~\citep{reliability_of_watermark} and computing $P_w(m|\mathbf{x}_{sliding\_window}).$\footnote{Applying Bayes' Theorem, we know that $P_w(m|\mathbf{x})$ is proportional to $P_w(\mathbf{x}|m)$, which can be computed as explained in Section~\ref{subsec: message extracting phase}. By normalizing $P_w(\mathbf{x}|m)$, we obtain $P_w(m|\mathbf{x})$.} To avoid incorrectly labeling human-written texts as watermarked texts, we set a threshold of $1-10^{-5}$. Any $\mathbf{x}_{sliding\_window}$ with $\max_m P(m|\mathbf{x})$ below this threshold will make it be classified as human-written. Our tests reveal that this threshold successfully prevents mislabeling human-written texts as watermarked. Figure~\ref{fig:cp} demonstrates the robustness of both Balance-Marking and Vanilla-Marking methods against copy-paste attacks, with Balance-Marking maintaining superior performance.

% not finished: add details to appendix

% \begin{figure}[t]
%     \centering
% \includegraphics[width=0.8\textwidth]{newfigs/lm-10-cprandom-10-cp.png}
%     \caption{The relationship between Success$_\text{m}$ after Copy-Paste Attacks and PPL. Balance-Marking outperforms Vanilla-Marking, especially under the coding rate of 5 tokens per bit.}
%     \label{fig:cp}
% \end{figure}

% not finished : need to add an appendix for substitution protocal
\paragraph{Robustness to Substitution Attacks.}
% In the practical application of model-generated texts, tokens may be replaced for editing purposes or to prevent watermark detection. To replicate this, we employ the Roberta-Large model to carry out word substitution. For a chosen text to be tested on, we arbitrarily pick an unaltered token each time. This token is then masked and the model is asked to predict it. In the event that a token's predicted logit surpasses the predicted logit of the original token minus 1.0, we replace the original token with the new one. With a designated substitution ratio $\alpha$ and a sentence consisting of $L$ tokens, we continue this process until the substituted tokens reach the value of $\alpha L$ or after $3\alpha L$ attempts are made. From empirical analysis, we establish that such replacements result in a marginal increase in PPL by around 0.1.

In the practical application of model-generated texts, tokens may be replaced for editing purposes or to prevent watermark detection. To replicate this, we employ the RoBERTa-Large~\citep{roberta} model to carry out word substitution (see details in Appendix~\ref{appendix: details_sub_attack}). %For a chosen text to be tested on, we arbitrarily pick an unaltered token each time. This token is then masked and the model is asked to predict it. In the event that a token's predicted logit surpasses the predicted logit of the original token minus 1.0, we replace the original token with the new one. With a designated substitution ratio $\alpha$ and a sentence consisting of $L$ tokens, we continue this process until the substituted tokens reach the value of $\alpha L$ or after $3\alpha L$ attempts are made. From empirical analysis, we establish that such replacements result in a marginal increase in PPL by around 0.1.
Figure~\ref{fig:sub-0.05} illustrates the impact of Substitution Attacks (substitution ratio=5\%) on Vanilla-Marking and Balance-Marking. Balance-Marking outperforms Vanilla-Marking as under Copy-Paste Attacks. Experiments with a substitution ratio of 10\% are reported in Appendix~\ref{appendix:sub-10}.

\subsection{Further Analysis}
\label{subsec: further analysis}
\begin{figure}[t]
    \centering
    \begin{subfigure}[b]{0.32\textwidth}  
       \includegraphics[width=\linewidth]{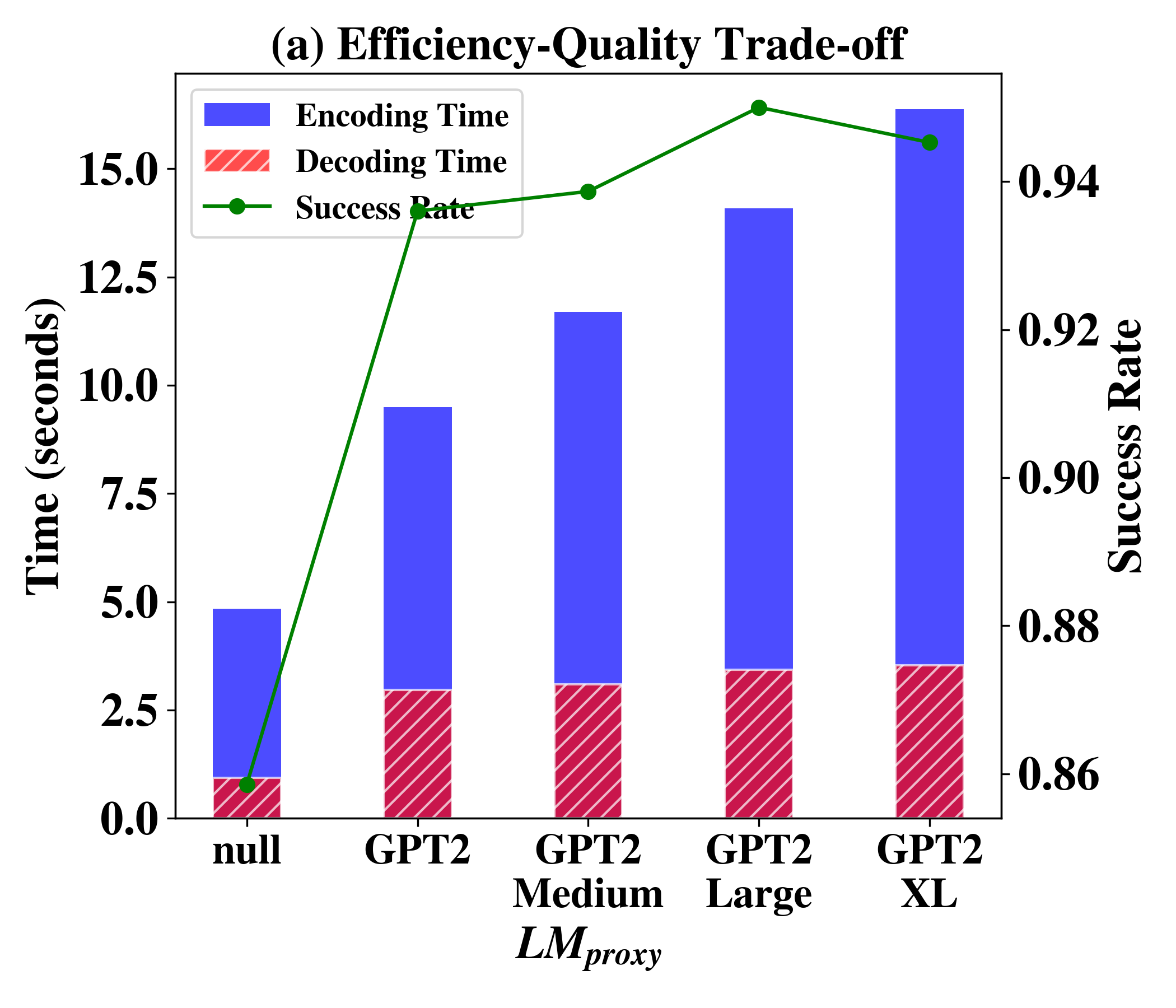}
    % \caption{Trade-off between efficiency and watermark success rate. Specifically, Vanilla-Marking can be viewed as Balance-Marking with $LM_{proxy} = \O$ (denoted as ``null'').}
    % \subcaption{}
    \phantomsubcaption
    \label{fig: trade_off_e_wq}
    \end{subfigure}
    \hfill
    \begin{subfigure}[b]{0.6\textwidth}
          \centering
\includegraphics[width=\linewidth]{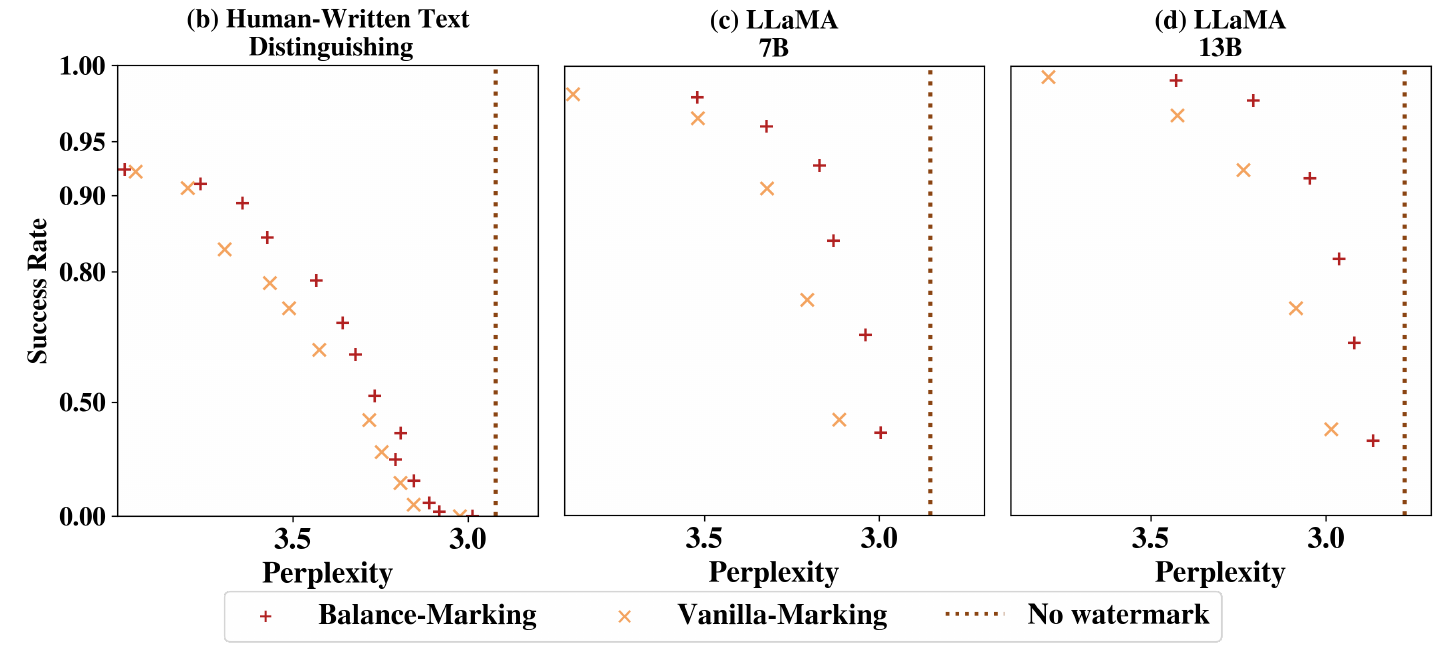}
    % \caption{Success$_\text{h}$ v.s. $1-10^{-5}$ PPL. Balance-Marking outperforms Vanilla-Marking.}
    % \subcaption{}
    \phantomsubcaption
    \label{fig:ppl-10-WSR-1-1e-5}
    \end{subfigure}
    \hfill
    \begin{subfigure}[b]{0pt}  
% \includegraphics[width=\linewidth]{newfigs/lm-7brandom-7b.png}
    % \caption{Balance-Marking outperforms Vanilla-Marking in larger LLMs (LLaMA-7B).}
   % \phantomsubcaption
   % \subcaption{}
   \phantomsubcaption
    \label{fig: scaling-7b-10}
     \end{subfigure}
    \hfill
    \begin{subfigure}[b]{0pt}  
% \includegraphics[width=\linewidth]{newfigs/lm-13brandom-13b.png}
    % \caption{Balance-Marking outperforms Vanilla-Marking in larger LLMs (LLaMA-13B).}
    % \phantomsubcaption
    % \subcaption{}
    \phantomsubcaption
    \label{fig: scaling-13b-10}
     \end{subfigure}
% \caption{Further Analysis on (a) human-written text distinguishing, (b) efficiency-quality trade-off, and (c) watermark on larger LLMs.}

\caption{(a) Trade-off between efficiency and watermark success rate. Specifically, Vanilla-Marking can be viewed as Balance-Marking with $LM_{proxy} = \O$ (denoted as ``null''). (b) Balance-Marking outperforms
Vanilla-Marking in distinguishing between message-embedded and human-written texts. (c, d) Balance-Marking is still superior to Vanilla-Marking when using LLaMA-7/13B.}
\label{fig: eefficiency and success rate}
\end{figure}

\paragraph{Distinguishing between message-embedded and human-written texts.}
\label{subsec: distinguishing}
% In the sections above, we measure the watermark quality by Success$_\text{m}$, success rate of restoring the injected watermark message. In practical scenarios, we may also need to avoid misclassifying human-written texts as watermarked texts and extracting messages from them. To achieve this, we can use a threshold to filter the human-written messages, as Section~\ref{sec: copy-paste attack} has done. Still, we use threshold  $1-10^{-5}$ as Section~\ref{sec: copy-paste attack}, which has been proven effective to filter all human-written texts. Figure~\ref{fig:ppl-10-WSR-1-1e-5} shows the Success$_\text{h}$ and text quality under different $\delta$s and coding rate 10 tokens per bit. Here, since all human-written text are classified correctly under threshold  $1-10^{-5}$, Success$_\text{h}$ actually measures the proportion of watermarked texts that misclassified as human-written. 
%Our previous assessment of watermark quality hinged on the Success$_{m}$ metric, which measures the success rate of retrieving the embedded watermark message. However, 
In practice, it's very important to ensure that human-written texts are not wrongly identified as watermarked. To prevent extracting pseudo messages from human-written texts, we use the same threshold $1-10^{-5}$ as in Section~\ref{sec: copy-paste attack}. While this threshold guarantees to prevent all human-written texts from being incorrectly identified, it may also misidentify some watermarked texts as human-written. Figure~\ref{fig:ppl-10-WSR-1-1e-5} demonstrates this by showing both the success rate of correctly distinguishing message-embedded texts from human-written ones, and the corresponding text quality (the coding rate is 10 tokens per bit). 
%It's worth noting that Figure~\ref{fig:ppl-10-WSR-1-1e-5} mirrors Figure~\ref{fig:cp} from the Copy-Paste Attack Section (Section~\ref{sec: copy-paste attack}), as both figures utilize the threshold. 
Still, Balance-Marking outperforms Vanilla-Marking.

% We previously evaluated watermark quality using the Success$_\text{m}$ metric, which gauges the success rate of restoring the injected watermark message. In real-world applications, however, it's just as crucial to prevent the misidentification of human-written texts as watermarked ones and avoid extracting false messages from them. To mitigate this issue, the threshold of $1-10^{-5}$ in Section~\ref{sec: copy-paste attack} can be utilized to filter out human-written messages. This threshold can effectively filter out all human-written texts, but some watermarked texts may be incorrectly identified as human-written. Figure~\ref{fig:ppl-10-WSR-1-1e-5} illustrates both the Success$\text{h}$, i.e. the success rate of recognizing the model-generated texts from human-written texts, and the corresponding text quality, under the coding rate of 10 tokens per bit. It's worth noting that Figure~\ref{fig:ppl-10-WSR-1-1e-5} bears similarities to Figure~\ref{fig:cp} from the Copy-Paste Attack section~(Section~\ref{sec: copy-paste attack}), given that they both involve a threshold of  $1-10^{-5}$.

\paragraph{Trade-off between Efficiency and Watermark Quality.}
% As stated in Section~\ref{sec: experiment}, Balance-Marking can generally outperform Vanilla CTWL, due to its utilization of $LM_{proxy}$. $LM_{proxy}$ can provide information of the probability of the preceding token, helping maximize $L(m,\mathbf x^{prompt}, \mathbf{t}_{:(l-1)})$ during the design of $P_{w}$\wang{may need to change the symbol P}. However, introducing $LM_{proxy}$ and the additional calculation of $V_{m,\mathbf{t}_{:(l-1)}}$ add up to the computational cost in the encoding and decoding procedure, as shown in Table~\ref{tab:main_efficieny}.
% Referring back to Section~\labelcref{sec: main_results,subsec: robustness_attack,subsec: distinguishing}, it is established that Balance-Marking generally outperforms Vanilla-Marking. This superior performance is attributable to the application of $LM_{proxy}$, which provides information about the probability of a token's predecessor, thereby aiding the maximization of $L(m,\mathbf x^{prompt}, \mathbf{t}_{:(l-1)})$ during the design of $P_{w}$. However, the integration of $LM_{proxy}$ and the additional computation of $V_{m,\mathbf{t}_{:(l-1)}}$ increase the computational costs of the encoding and decoding process, as reflected in Table~\ref{tab:main_efficieny}.
The $LM_{proxy}$ introduced in Balance-Marking brings better watermark quality but also along with extra computational cost. In Figure~\ref{fig: trade_off_e_wq}, we verify this trade-off between the efficiency and watermark quality when larger $LM_{proxy}$s are introduced. Specifically, we can view Vanilla-Marking as a special kind of Balance-Marking with $LM_{proxy} = \O$. We also discussed the $LM_{proxy}=LLM$ case in Appendix~\ref{app:anothertable}

\paragraph{Scaling to larger LLMs.}
\label{para:scaling}
To verify that our watermark algorithm also works for larger LLMs, we conduct experiments with LLaMA-7/13B (under a coding rate of 10, and a smaller $\delta$ set $\{0.8, 1.0, 1.2, 1.5, 2.0, 3.0\}$ to save computational costs), as shown in Figure~\ref{fig: scaling-13b-10}. 
% Perplexity is calculated by LLaMA-33B. 
Appendix~\ref{appendix: scaling} also lists the results with a coding rate of 5. Balance-Marking stably outperforms Vanilla-Marking when scaling to larger LLMs.
% \begin{figure}[t]
%     \centering
% \includegraphics[width=0.8\textwidth]{newfigs/lm-13brandom-13b.png}
%     \caption{Balance-Marking outperforms Vanilla-Marking in larger LLMs (LLaMA-13B).}
%     \label{fig: scaling-13b-10}
% \end{figure}

% remove 6.2-6.5 to appendix, 6.1 to section 5

% 先注释掉，之后再看看有没有地方补回来
% \section{Limitations and Future Work}
% \label{sec: limitation}
% This is an ongoing iterative work, and we summarize the limitations of our current version and discuss future work in this section. 
% (1) Firstly, the sizes of our experimental base models do not cover all model scales, so the conclusions obtained may be affected by the scaling law; we will supplement more experimental results on models of different sizes and structures in future versions. 
% (2) Secondly, the datasets we evaluated were limited to the natural text domain. We will explore and construct more diverse watermarking evaluation benchmarks, including code segments, logic problems, and more fine-grained natural texts in future versions. 
% (3) Furthermore, our method performs poorly when facing paraphrase attacks, which is actually a significant threat to all watermarking methods. We look forward to developing watermarking algorithms that can resist paraphrase attacks. 
% (4) In addition, we will continue to focus on and improve the vital trade-offs of watermarking algorithms, such as effectiveness and efficiency, and coding rate and robustness.

\section{Conclusion}
\label{sec: conclusion}
In this work, we provide the first systematic study on the topic of Codable Text Watermarking for Large Language Models (\textbf{CTWL}), filling the research gap of integrating multi-bit information watermarks into the generation process of LLMs. 
We first conduct a taxonomic analysis of CTWL and provide a rigorous mathematical formalization for CTWL. 
We then design a CTWL method Balance-Marking, which effectively ensures the balance of the probabilities of available and unavailable vocabularies with the help of a proxy language model, thereby ensuring the high quality and diversity of the generated text. 
Extensive experiments have shown that our method significantly outperforms the direct baseline method in the comprehensive evaluation of five dimensions, reaching a practical level of usability. 
We hope that our work can help the community better understand the CTWL issue and inspire more outstanding research in the future.

% Our contributions mainly lie in three aspects.
% (1) 
% %From a research perspective, 
% We conduct a taxonomic analysis of CTWL and provide a rigorous mathematical formalization for CTWL. 
% (2) 
% %From an application perspective, 
% We analyze the demand for CTWL in diverse application scenarios of LLMs and summarize a comprehensive evaluation system for the application of CTWL from 5 evaluation dimensions. 
% (3) 
%From an empirical perspective, 

%without significantly increasing computational complexity. 

%Further analysis has revealed the underlying mechanisms of the various modules in our method.

\section*{Ethical Statement}

Our paper aims to provide an effective way to reduce the ethical problems brought by the LLMs. Our method can inject a watermark that carries rich information into the text generated by the LLMs, thus helping the public to identify whether the specific text is machine-generated under some necessary circumstances. For example, a machine-generated text with watermarks that carry the information of the name of the source LLM can be easily traced from the source once it is used for harmful purposes, such as creating fake news or cheating on academic writings. Therefore, our work does not have any negative ethical concerns. 

\section*{Reproducibility Statement}
We have provided the necessary explanations for all the assumptions we made after they are proposed in the main paper.  We provide detailed mathematical derivations for the complicated formulas such as deriving Eq.~(\ref{eq: encoding message and text quality original}) to Eq.~(\ref{eq: encoding optimization problem}) in Appendix~\ref{appendix: math derivations}. We present the detailed illustrations of the approximations we made during mathematical derivations in Appendix~\ref{appendix: reasons for omitting}. We give the necessary illustration of the experiment settings in Section~\ref{subsec: experimental settings} and Section~\ref{subsec: experimental settings} in the main paper. The detailed explanations and explorations of training hyper-parameters are in Appendix~\ref{appendix: hyperprameters}. The descriptions of the implementations of substitution attacks are in Appendix~\ref{appendix: details_sub_attack}.

\section*{Acknowledgements}
We sincerely thank all the anonymous reviewers for their helpful suggestions on improving our manuscript. This work is supported in part
by a Tencent Research Grant and the National Natural
Science Foundation of China (No. 62176002). Yankai Lin and Xu Sun are the corresponding authors of this paper.

\bibliography{iclr2024_conference}
\bibliographystyle{iclr2024_conference}

\newpage

\appendix

\section{Supplement to Related Work}
\label{appendix:detection_method}
\paragraph{Detecting-based Methods.} This kind of method aims to detect whether a text is generated by a language model by formulating the task as a binary classification problem~\citep{real_or_fake,automatic_detection,gpt-2-release}. (1) One way to achieve this goal is to first collect outputs from a specific LLM along with human-written texts, and then use them to train a binary classifier~\cite{chatgpt_or_human}. For example,~\citet{gpt-2-release} fine-tune a RoBERTa classification model to detect whether a text is generated by GPT-2 model or not. 
Later,~\cite{chatgpt_or_human} collect data produced by ChatGPT~\cite{chatgpt} and fine-tune a DistilBERT model~\citep{distilbert} to distinguish texts written by humans and that generated by ChatGPT.
~\citet{openai_classifier} create a more advanced classifier that can detect texts generated by a variety of diverse LLMs. 
(2) Besides explicitly training deep classifiers, some other methods achieve detection by exploring the statistical differences between LLM-generated texts and human-written texts~\citep{release_strategies,llmdet}. 
GPTZero~\citep{gptzero} is a popular tool to identify machine-generated texts by calculating the perplexity and burstiness scores of texts and comparing them with specific thresholds.
DetectGPT~\citep{detectgpt} claims that when a machine-generated text is perturbed, its log probability will always show a decreasing pattern, which is not the same for human-written texts. 
Based on this finding, it defines a log probability-based metric to help detect texts generated by LLMs. 
Recently, there have been some works~\citep{new-detection-1, new-detection-2} exploring the detection of machine-generated text in specific domains.
However, all these detection methods face the risk of becoming increasingly ineffective in the future, as the LLMs will be consistently improved to behave more and more like humans, which finally makes detection impractical~\citep{reliably_detect}.

\section{Detailed mathematical derivations omitted in the main paper}
\label{appendix: math derivations}
Here, we provide detailed mathematical derivations about the formulas that are simplified in the main paper.

\paragraph{Eq.~(\ref{eq: target on encoding messages}):}

\begin{equation}
%\label{eq: target on encoding messages}
\begin{aligned}
 & \mathop{\max}\limits_{\mathbf{t}}  \{ P_{w}(\mathbf{t}|m) / 
 \mathop{\max}_{m' \neq m} P_{w}(\mathbf{t}|m')  \}
 \\ &\iff
  \mathop{\max}\limits_{\mathbf{t}} \{ \mathop{\Pi}\limits_{l=1}^{L} P_{w}(t_{l}|,m,\mathbf{t}_{:(l-1)}) / \mathop{\max}_{m' \neq m} \mathop{\Pi}\limits_{l=1}^{L} P_{w}(t_{l}|m',\mathbf{t}_{:(l-1)})  \} 
 \\ &\iff
 \mathop{\max}\limits_{\mathbf{t}} \{ \mathop{\sum}\limits_{l=1}^{L} \mathop{\log}P_{w}(t_{l}|m,\mathbf{t}_{:(l-1)}) - \mathop{\max}_{m' \neq m} \mathop{\sum}\limits_{l=1}^{L} \mathop{\log}P_{w}(t_{l}|m',\mathbf{t}_{:(l-1)})  \} .
\end{aligned}
\end{equation}

\paragraph{From Eq.~(\ref{eq: encoding message and text quality original}) to Eq.~(\ref{eq: encoding optimization problem}):}
As $\text{PPL}(\mathbf{t}|\mathbf{x}^{prompt})=[\mathop{\Pi}\nolimits_{l=1}^{L}P(t_{l}|\mathbf{x}^{prompt},\mathbf{t}_{:(l-1)})]^{-\frac{1}{L}} $, Eq.~(\ref{eq: encoding message and text quality original}) is equivalent to 
\begin{equation}
\begin{aligned}
& \mathop{\max}\limits_{\mathbf{t}} \{ \mathop{\sum}\limits_{l=1}^{L} \mathop{\log}P_{w}(t_{l}|m,\mathbf{t}_{:(l-1)}) - \mathop{\max}_{m' \neq m} \mathop{\sum}\limits_{l=1}^{L} \mathop{\log}P_{w}(t_{l}|m',\mathbf{t}_{:(l-1)})  \} , 
 \\
& \text{s.t. \quad } -\frac{1}{L}[\mathop{\sum}\limits_{l=1}^{L} \log P(t_{l}|\mathbf{x}^{prompt},\mathbf{t}_{:(l-1)})] \leq %-\frac{1}{L^{ori}}[\mathop{\Pi}\limits_{l=1}^{L^{ori}}P(t_{l}^{ori}|\mathbf{x}^{prompt},\mathbf{t}_{:(l-1)}^{ori})] +\epsilon .
\log (\text{PPL}(\mathbf{t}^{ori}| \mathbf{x}^{prompt})  +\epsilon ).
\end{aligned}
\end{equation}
By applying the method of Lagrange Multipliers, we can re-define the target as:
\begin{equation}
\label{eq: encoding message and text quality method 2}
\begin{aligned}
\mathop{\max}_{\mathbf{t}} \{& \mathop{\sum}\limits_{l=1}^{L} \mathop{\log}P_{w}(t_{l}|m,\mathbf{t}_{:(l-1)}) -\mathop{\max}_{m' \neq m} \mathop{\sum}\limits_{l=1}^{L} \mathop{\log}P_{w}(t_{l}|m',\mathbf{t}_{:(l-1)})\\ & -\lambda[
% \text{PPL}(\mathbf{t}|\mathbf{x}^{prompt}) - \min\limits_{\mathbf{t}^{'}} \text{PPL}(\mathbf{t}^{'}| \mathbf{x}^{prompt}) -\epsilon
-\frac{1}{L}[\mathop{\sum}\limits_{l=1}^{L} \log P(t_{l}|\mathbf{x}^{prompt},\mathbf{t}_{:(l-1)})] -
\log (\text{PPL}(\mathbf{t}^{ori}| \mathbf{x}^{prompt})  +\epsilon )
] \}.
% & \text{s.t. \quad } \text{PPL}(t|\mathbf{x}^{prompt},m) \leq \min\limits_x\text{PPL}(x| \mathbf{x}^{prompt}) +\epsilon ,
\end{aligned}
\end{equation}
For simplicity but without loss of generality, let's assume that the PPL scores are calculated based on the same $LLM$ used to generate $t$.\footnote{This assumption is practical as the LLM is powerful enough to accurately measure the perplexity of a text.} In this case, given the prompt $\mathbf{x}^{prompt}$ and $LLM$, the term $ \text{PPL}_{LLM}(\mathbf{t}^{ori}| \mathbf{x}^{prompt})$ can be regarded as a constant. 
Therefore, Eq.~(\ref{eq: encoding message and text quality method 2}) can be rephrased as:
\begin{equation}
\begin{aligned}
\mathop{\max}_{\mathbf{t}} \{& \mathop{\sum}\limits_{l=1}^{L} \mathop{\log}P_{w}(t_{l}|m,\mathbf{t}_{:(l-1)}) -\mathop{\max}_{m' \neq m} \mathop{\sum}\limits_{l=1}^{L} \mathop{\log}P_{w}(t_{l}|m',\mathbf{t}_{:(l-1)})\\
&+\frac{\lambda}{L} \mathop{\sum}\limits_{l=1}^{L} \mathop{\log}P_{LLM}(t_{l}|\mathbf{x}^{prompt},\mathbf{t}_{:(l-1)}) \}.
% & \text{s.t. \quad } \text{PPL}(t|\mathbf{x}^{prompt},m) \leq \min\limits_x\text{PPL}(x| \mathbf{x}^{prompt}) +\epsilon ,
\end{aligned}
\end{equation}
Let $\delta=\frac{L}{\lambda}$ and $\hat m=\mathop{\mathrm{argmax}}\limits_{m' \neq m} \mathop{\sum}\limits_{l=1}^{L} \mathop{\log}P(t_{l}|m',\mathbf{t}_{:(l-1)})$, the objective function can be restructured as Eq.~(\ref{eq: encoding optimization problem}).

\section{The Evaluation System of Codable LLM Watermarking}
\label{sec: evaluation}
The prosperity of LLM technology has brought more diverse and differentiated application scenarios than traditional NLP models. 
Therefore, the evaluation of the practicability of LLM-generated text watermarking technology is expected to keep pace with this trend.
However, existing studies lack a unified convincing evaluation system for LLM-generated text watermarking technology.
In this section, we start with analyzing the uniqueness of LLM applications and establish a comprehensive evaluation system for LLM-generated text watermarking from 5 different aspects.
\subsection{Watermarking Success Rate}
We define two indicators to measure how successful the watermark is injected into LLM-generated texts as expected: (1) Success rate of recognizing the model-generated texts from human written texts, and (2) Success rate of recovering the injected watermark message.
Both of these two metrics need to be considered for CTWL; while for normal LLM watermarking~\citep{watermark_llm}, only the first metric need to be evaluated. 

\subsection{Robustness Against Attacks}
Texts generated by LLMs are usually modified before they are actually used for the purpose of polishing or detection escaping. Thus, the watermarking algorithms need to ensure robustness in the face of various corruptions and attacks.
We summarize the two most representative attacks threatening the success of LLM watermarking: (1)~\textbf{Copy-Paste Attack}~\citep{reliability_of_watermark}, where LLM-generated text fragments are mixed with human-written text fragments; (2)~\textbf{Substitution Attack}, where individual or sequential tokens are synonymously replaced based on human knowledge or masked language models like BERT~\citep{bert} or RoBERTa~\citep{roberta}.

\subsection{Payload Information Coding Rate}
For non-codable watermarking methods, the encoded information is always one-bit. For CTWL, we divide the number of bits carried by the watermark by the length of the covered tokens as the indicator for measuring the coding rate of the watermarking algorithm.
Obviously, a good watermarking technique should encode as many bits of information as possible without sacrificing the performance on other metrics.

\subsection{Encoding and Decoding Efficiency}
Injecting watermarks during LLM's generation will inevitably increase the computational cost. Moreover, restoring multiple bits of information from the text also takes higher computational complexity than decoding 1-bit information. We argue that it is necessary to consider the additional computational complexity brought by the encoding and decoding of the large model watermarking algorithm. 
Besides, the parallelism of the watermarking algorithm is also vital for the actual time consumption.

\subsection{Impact on the Quality of Generated Text}
Codable watermarks contain more complex information and have a larger impact on the quality of text generated by LLMs than non-codable watermarks. Therefore, it is necessary to ensure that the impact of the watermark on the quality of LLM-generated text is within an acceptable range for the deployment of watermarking algorithms. In the current work, we adopt the text perplexity (PPL) as an automated metric to measure the quality of LLM-generated texts. In future work, we will include more metrics such as semantic similarity or human evaluation for comprehensive evaluation of text quality.
%As an alternative supplementary, manually reviewing the LLM-generated text with watermarks can further confirm the impact of watermark on the quality of texts.
% V2 ToDo
% \subsection{Datasets}

\section{Potential Application Scenes of Codable LLM Watermarking}
\label{sec: scenes}
In this section, we analyze some potential application scenarios of codable LLM watermarking, and examine how our method adjusts the proxy-LM to adapt to the varying demands for watermarking techniques in different application scenarios.

\subsection{Corporate Intellectual Property Protection}
For service providers based on LLMs, generating texts with watermarks containing information related to the model, service, request, or user can effectively ensure that the generated texts can be traced and identified, thereby preventing the model from being used unreasonably or without authorization. Service providers can flexibly choose and combine the information to be included in the watermark to better protect their intellectual property. 
In this scenario, as the owner of LLMs, the service provider can choose to use the large model itself as a proxy-LM to minimize the impact of adding watermarks on the quality of the text, or prepare a small model similar to the large one to accelerate inference through methods like distillation, quantization, or pruning.

\subsection{User-level Copyright Protection}
The discussion on the copyright of text generated by LLMs is also an interesting topic, as users may believe that their intellectual input when writing prompts gives them (at least part of) the copyright of the generated text. In such a situation, users can choose to reach an agreement with the LLMs service provider on a customized watermark algorithm for the user self (through customizing the proxy-LM or hash function). In the encoding stage, the service provider will generate text containing a specific watermark according to the user's exclusive watermark encoding algorithm. When the user wants to prove that a piece of text comes from him- or herself, the user can request the service provider to use his or her exclusive decoding algorithm to confirm whether it contains a personal watermark. To make this more credible, an independent third-party organization similar to the patent office can take the responsibility to manage and certify these customized watermark algorithms.

\subsection{Open Watermarking Protocol}
For the public, it is desirable to have a very convenient way to identify whether a text comes from a model and which model it comes from. 
We propose an idea based on an open watermark protocol to reduce the identification problems caused by scattered and differentiated watermarking methods adopted by different service providers. 
First, the protocol selects an open-source language model (such as GPT-2) as a proxy-LM, and then determines a unified and scalable message coding system to establish a publicly available watermark encoding and decoding algorithm. 
Any service provider that joins this protocol can use the watermark encoding algorithm to inject watermarks into the texts generated by their private models by extending the message coding system. 
In this way, the public can efficiently identify all models that join the protocol using a single decoding algorithm. 
The technical support for this idea to work is that our Balance-Marking leverages both the LLMs and the proxy-LM during the encoding stage, while in the decoding stage, only the proxy-LM is needed to restore the watermark information. This makes it possible to have multiple closed-source LLMs encoding and a single public model decoding.
If as many service providers as possible join this protocol, identifying the source of the text will become an increasingly easy task, which can effectively alleviate the impact of LLM-generated texts on human community order and security.

\subsection{Relay Watermarking among Models}
Model-generated text that is actually applied is usually not generated in one step, but may go through multiple users or multiple models for processing, such as expansion, polishing, translation, paraphrasing, and so on. If we want to track the complete production process of a text, not just obtain the information of the last model processing, it is feasible to ensure that the watermark is incrementally written in a relay form among different models. As discussed in the previous subsection, this requires an open watermark protocol and a scalable information encoding system. This also requires adding a processing step to our method: first extract the watermark message from the text input to the model, mix it with the message of the new model itself, and then add the new message to the watermark of the newly generated text. In this way, we can track the complete life cycle of machine-generated text. Of course, as discussed in the previous subsection, this also requires as many LLM service providers as possible to join the open watermark protocol.

\section{Approximation during calculation $P_w$}
\label{appendix: reasons for omitting}

In calculating $P_w$ of Vanilla-Marking and Balance-Marking, we perform the following two approximations. \paragraph{Excluding Term $\frac{1}{|\mathcal{M}|}\mathop{\sum}\nolimits_{m'\in \mathcal{M}}\mathop{\log}P_{w}(v|m',\mathbf{t}_{:(l-1)})$ from Encoding} When implementing Algorithm~\ref{algo: encoding algorithm for general P}, there is a need to compute $\frac{1}{|\mathcal{M}|}\mathop{\sum}\nolimits_{m'\in \mathcal{M}}\mathop{\log}P_{w}(v|m',\mathbf{t}_{:(l-1)})$, which can be a time-consuming task. However, due to the inherent randomness of $P_w$, we hypothesize that this mean value almost remains constant under the Law of Large Numbers. This hypothesis is corroborated by our empirical studies~(see Appendix~\ref{appendix:omit1}) that indicate that this value remains close to -11 with a relatively insignificant standard deviation (less than 0.05) across $v$. Hence, we can exclude this calculation from Algorithm~\ref{algo: encoding algorithm for general P} during encoding for efficiency.
\paragraph{Substituting $ \log P_w(v|m,\mathbf{t}_{:(l-1)})$ with $\log \hat{P}_w(v|m,\mathbf{t}_{:(l-1)})$ during Decoding} Similarly, we \textbf{substitute $ \log P_w(v|m,\mathbf{t}_{:(l-1)})$ with $\log \hat{P}_w(v|m,\mathbf{t}_{:(l-1)})$ in Eq.~(\ref{eq: decoding messages})}. The difference between the two expressions equals $\mathop{\log}\mathop{\sum}\limits_v \hat{P}_w(v|m,\mathbf{t}_{:(l-1)})$, which we assume to be almost constant. Our empirical findings~(see Appendix~\ref{appendix:omit2}) support this claim. In the Vanilla-Marking approach, the standard deviation is as low as 0.002, with an average value of 11.4. Despite Balance-Marking having a slightly higher deviation (0.23) and a mean of 11.4, it still allows for the exclusion of this calculation to help speed up the computation process.

\subsection{Excluding Term $\frac{1}{|\mathcal{M}|}\mathop{\sum}\nolimits_{m'\in \mathcal{M}}\mathop{\log}P_{w}(v|m',\mathbf{t}_{:(l-1)})$ from encoding}
\label{appendix:omit1}
The randomness present within the design of the message function $P_w$ allows us to interpret expressions of $\log P_{w}(v|m',\mathbf{t}_{:(l-1)})$s in $\frac{1}{|\mathcal{M}|}\mathop{\sum}\nolimits_{m'\in \mathcal{M}}\mathop{\log}P_{w}(v|m',\mathbf{t}_{:(l-1)})$ as independent and identically distributed (i.i.d.) random variables. Consequently, via the law of large numbers, the term $\frac{1}{|\mathcal{M}|}\mathop{\sum}\nolimits_{m'\in \mathcal{M}}\mathop{\log}P_{w}(v|m',\mathbf{t}_{:(l-1)})$ remains approximately constant.

As a practical test of our hypothesis, we randomly selected 100 sequences of $\mathbf{t}_{:(l-1)}$ from human-written texts (specifically, the news-like subset of the C4 dataset~\citep{2019t5}) as well as watermarked texts. Afterward, for each $\mathbf{t}_{:(l-1)}$, we randomly picked 100 tokens $v$ from the vocabulary to calculate the standard deviation of $\frac{1}{|\mathcal{M}|}\mathop{\sum}\limits_{m'\in \mathcal{M}}\log P_{w}(v|m',\mathbf{t}_{:(l-1)})$ across $v$.\footnote{Since the solution $t_l$ depends on the model logit and message logit of $v$ under a given $\mathbf{t}_{:(l-1)}$, it's enough to concentrate on the variations across $v$.}
%\footnote{$\mathop{\log}\mathop{\sum}\limits_v \hat{P}_w(v|m,\mathbf{t}_{:(l-1)})$ in $\frac{1}{|\mathcal{M}|}\mathop{\sum}\nolimits_{m'\in \mathcal{M}}\mathop{\log}P_{w}(t_{l}|m',\mathbf{t}_{:(l-1)})$ for Vanilla-Marking is difficult to compute, since we have to compute $\mathop{\log}\mathop{\sum}\limits_v \hat{P}_w(v|m,\mathbf{t}_{:(l-1)})$ for $|\mathcal{M}|$ times. Since Appendix~\ref{appendix:omit2} shows that $\mathop{\log}\mathop{\sum}\limits_v \hat{P}_w(v|m,\mathbf{t}_{:(l-1)})$ is nearly a constant, when  we use the mean of $\mathop{\log}\mathop{\sum}\limits_v \hat{P}_w(v|m,\mathbf{t}_{:(l-1)})$ in Eq.~(\ref{eq:18}), instead of calculating $\mathop{\log}\mathop{\sum}\limits_v \hat{P}_w(v|m,\mathbf{t}_{:(l-1)})$ for each $m$. For the standard deviation, we use the sum of the deviation of $\frac{1}{|\mathcal{M}|}\mathop{\sum}\nolimits_{m'\in \mathcal{M}}\mathop{\log}\hat P_{w}(t_{l}|m',\mathbf{t}_{:(l-1)})$ and the deviation of $\mathop{\log}\mathop{\sum}\limits_v \hat{P}_w(v|m,\mathbf{t}_{:(l-1)})$ as an estimation.} 
We report the standard deviation and mean across $v$ (averaged over 100 instances of $\mathbf{t}_{:(l-1)}$s) in Table~\ref{tab:omit1}.  Given the small standard deviation across the vocabulary, it's appropriate to exclude it in Algorithm~\ref{algo: encoding algorithm for general P}.

\begin{table}[t]
    \centering
    \caption{$\frac{1}
{|\mathcal{M}|}\mathop{\sum}\limits_{m'\in \mathcal{M}}\mathop{\log}P_{w}(v|m',\mathbf{t}_{:(l-1)})$ on human-written texts and watermarked texts. The numbers are the mean $\pm$ and the standard deviation across $v$.}
    \begin{tabular}{c|c|c}
        \toprule
        &\textbf{Human-Written Texts} & \textbf{Watermarked Texts} \\
        \midrule
        Vanilla-Marking & -10.95 $\pm$ 0.0005 &  -10.95 $\pm$ 0.0005 \\
        \midrule
        Balance-Marking & -10.93 $\pm$ 0.0461 & -10.93 $\pm$ 0.0457 \\
        \bottomrule
    \end{tabular}
    \label{tab:omit1}
\end{table}

\subsection{substituting $ \log P_w(v|m,\mathbf{t}_{:(l-1)})$ with $\log \hat{P}_w(v|m,\mathbf{t}_{:(l-1)})$ during decoding}
\label{appendix:omit2}
Consider that 
\begin{equation}
\label{eq:18}
    \log P_w(v|m,\mathbf{t}_{:(l-1)}) = \log \hat P_w(v|m,\mathbf{t}_{:(l-1)}) - \mathop{\log}\mathop{\sum}\limits_v \hat{P}_w(v|m,\mathbf{t}_{:(l-1)}),
\end{equation}
in a manner akin to Appendix~\ref{appendix:omit1}, we hypothesize that $\mathop{\log}\mathop{\sum}\limits_v \hat{P}_w(v|m,\mathbf{t}_{:(l-1)})$ may remain nearly constant. If so, the magnitude of $P_w(v|m,\mathbf{t}_{:(l-1)})$ can be represented by the term $\hat{P}_w(v|m,\mathbf{t}_{:(l-1)})$. 

As Appendix~\ref{appendix:omit1}, the term $\log \hat P_w(v|m,\mathbf{t}_{:(l-1)})$ can be interpreted as i.i.d. random variables, hence by invoking the law of large numbers, for any $\epsilon$, we establish:
\begin{equation}
   \label{eq:19}
    \lim _{|\mathcal{V}| \rightarrow \infty} P\left(\left|\frac{\mathop{\sum}\limits_v \hat{P}_w(v|m,\mathbf{t}_{:(l-1)})}{|\mathcal{V}|}-\mu\right|>\varepsilon\right)=0,
\end{equation}
with $\mathcal{V}$ representing the vocabulary and $\mu$ is the mean.

According to the mean value theorem, a $c$ exists between $\mathop{\sum}\limits_v \hat{P}_w(v|m,\mathbf{t}_{:(l-1)})$ and $\mu|\mathcal{V}|$ such that
\begin{equation}
    \frac{\mathop{\log}\mathop{\sum}\limits_v \hat{P}_w(v|m,\mathbf{t}_{:(l-1)})  - \log (\mu|\mathcal{V}|)}{\mathop{\sum}\limits_v \hat{P}_w(v|m,\mathbf{t}_{:(l-1)}) - \mu |\mathcal{V}|} = \frac{1}{c} .
\end{equation}

Through empirical experimentation,\footnote{We use the same experimental settings as the experiments shown in Table~\ref{tab:omit2}.} it is observed that $\log\mathop{\sum}\limits_v \hat{P}_w(v|m,\mathbf{t}_{:(l-1)})$ exceeds 10.5. Given that the size of vocabulary $|\mathcal{V}|$ is 50257, it implies $c>e^{10.5}>  \frac{1}{2}|\mathcal{V}|$, leading to
\begin{equation}
\label{eq:21}
    |\mathop{\log}\mathop{\sum}\limits_v \hat{P}_w(v|m,\mathbf{t}_{:(l-1)})  - \log \mu|\mathcal{V}||< 2|\mathop{\sum}\limits_v \frac{\hat{P}_w(v|m,\mathbf{t}_{:(l-1)}}{|\mathcal{V}|}) - \mu|.
\end{equation}

Merging Eq.~(\ref{eq:19}) and Eq.~(\ref{eq:21}) confirms that, with high certainty, , $\mathop{\log}\mathop{\sum}\limits_v \hat{P}_w(v|m,\mathbf{t}_{:(l-1)})$ approximates a constant.

To validate our hypothesis, we  randomly select 100 $\mathbf{t}_{:(l-1)}$ as above, and randomly pick 100 $m$ from $\mathcal{M}$. The standard deviation and mean of $\mathop{\log}\mathop{\sum}\limits_v \hat{P}_w(v|m,\mathbf{t}_{:(l-1)})$  are presented in Table~\ref{tab:omit2}.
The deviation of $\mathop{\log}\mathop{\sum}\limits_v \hat{P}_w(v|m,\mathbf{t}_{:(l-1)})$ is higher in Balance-Marking than in Vanilla-Marking, due to the likelihood that  $\log \hat P_w(v|m,\mathbf{t}_{:(l-1)})$ may not be precisely i.i.d for Balance-Marking. Nevertheless, this modest variation still allows us to exclude $\mathop{\log}\mathop{\sum}\limits_v \hat{P}_w(v|m,\mathbf{t}_{:(l-1)})$ for the sake of swifter computation.

\begin{table}[t]
    \centering
    \caption{$\mathop{\log}\mathop{\sum}\limits_v \hat{P}_w(v|m,\mathbf{t}_{:(l-1)})$ on human-written texts and watermarked texts. The numbers are the average $\pm$ the standard deviation across $(\mathbf{t}_{:(l-1)}, m)$ .}
    \begin{tabular}{c|c|c|c}
        \toprule
        &\textbf{Human-Written Texts} & \textbf{Watermarked Texts} \\
        \midrule
        Vanilla-Marking & 11.45 $\pm$ 0.0019 & 11.45 $\pm$ 0.0022  \\
        \midrule
        Balance-Marking & 11.42 $\pm$ 0.1797 &  11.41 $\pm$ 0.2266 \\
        \bottomrule
    \end{tabular}
    \label{tab:omit2}
\end{table}

\section{Detailed illustrations about the practical improvements on balance-marking}
\label{appendix: improvements of balance-marking}
The method we introduced at the beginning of Section~\ref{subsubsec: balance-marking} may encounter several obstacles in realistic applications. Thus, we make the following improvements to make it more practical when facing various realistic circumstances:

\paragraph{(1) Omit $\mathbf{x}^{prompt}$ and truncate $\mathbf{t}_{:(l-1)}$ into a fixed length.} 
Considering that the $\mathbf{x}^{prompt}$ is usually unavailable and the text receiver might only obtains a segment of the watermarked text during the message decoding phase, 
it will cause inconsistency between the encoding and decoding phases on calculating the token probabilities for creating $V_{m,\mathbf{t}_{:(l-1)}}$. To address this problem, we employ $P_{LLM}(v| \mathbf{t}_{(l-1- L_{prefix}):(l-1)})$ to approximate $P_{LLM}(v|\mathbf{x}^{prompt},\mathbf{t}_{:(l-1)})$. That is, 
we omit $\mathbf{x}^{prompt}$ and truncate $\mathbf{t}_{:(l-1)}$ to a fixed-length $\mathbf{t}_{(l-1- L_{prefix}):(l-1)}$ for consistency during both encoding and decoding.

\paragraph{(2) Use a proxy language model (proxy-LM) $LM_{proxy}$ in Eq.~(\ref{eq: V_m condition}).} 
Moreover, considering several usage scenarios discussed in Appendix~\ref{sec: scenes}, 
we broaden the $LLM$ used in $P_{LLM}(v| \mathbf{t}_{(l-1- L_{prefix}):(l-1)})$ into a general proxy model denoted as $LM_{proxy}$. 
This model can either be $LLM$ itself when the model company wants the watermarked text only to be decoded by itself, or be another smaller and public language model (e.g., GPT-2~\citep{gpt2}) $P_{LLM}$ that allows for quicker computation of $P_{w}$ and enables anyone to decode the embedded message without knowing the specific $LLM$ used in text generation.

\paragraph{(3) Pre-map message space into a smaller space for efficient computing.} 
Since computing $V_{m,\mathbf{t}_{:(l-1)}}$ for each $m$ during encoding can be much time-consuming when the message space is pretty large, 
we opt to first map the entire message space into a smaller space as $m \rightarrow \hat{h}(m) \in \mathcal{M}_{A}=\{1,\cdots, A \}$ by using another hash function $\hat{h}$, and then compute the seed $s$ as $s = h(\hat h(m), x_{:(l-1)})$. 
By this way, we only need to run Algorithm~\ref{algo: choose V_{m,x_{:(l-1)}}_old} a mere $A$ times for each $\mathbf{t}_{:(l-1)}$.

\section{Hyperparameters and Hash Scheme}

\label{appendix: hyperprameters}
In Section~\ref{subsec: experimental settings}, we opt to use $LM_{proxy}= \mathrm{GPT2}$~\citep{gpt2}, $A=100$, $L_{prefix}=10$, $\sigma=0.5$ and $\mathcal{M}=\{0,1,...,2^{20}-1\}$ for Balance-Marking. The reason to set $\sigma=0.5$ is to achieve the maximal diversity of $V_{m, \mathbf{t}_{:(l-1)}}$ w.r.t. different $m$. The choice of $LM_{proxy}$, $A$, $L_{prefix}$, $\mathcal{M}$ aims to balance efficiency and performance, which is further discussed in Appendix~\labelcref{sec: ablation_lm,sec: The effect of $A$,sec: The impact of L_prefix,sec: The impact of |M|}. The hash scheme is detailed in Appendix~\ref{appsec: hash}

\subsection{The impact of the proxy language model $LM_{proxy}$}
\label{sec: ablation_lm}
\begin{figure}[t]
    \centering
\includegraphics[width=0.8\textwidth]{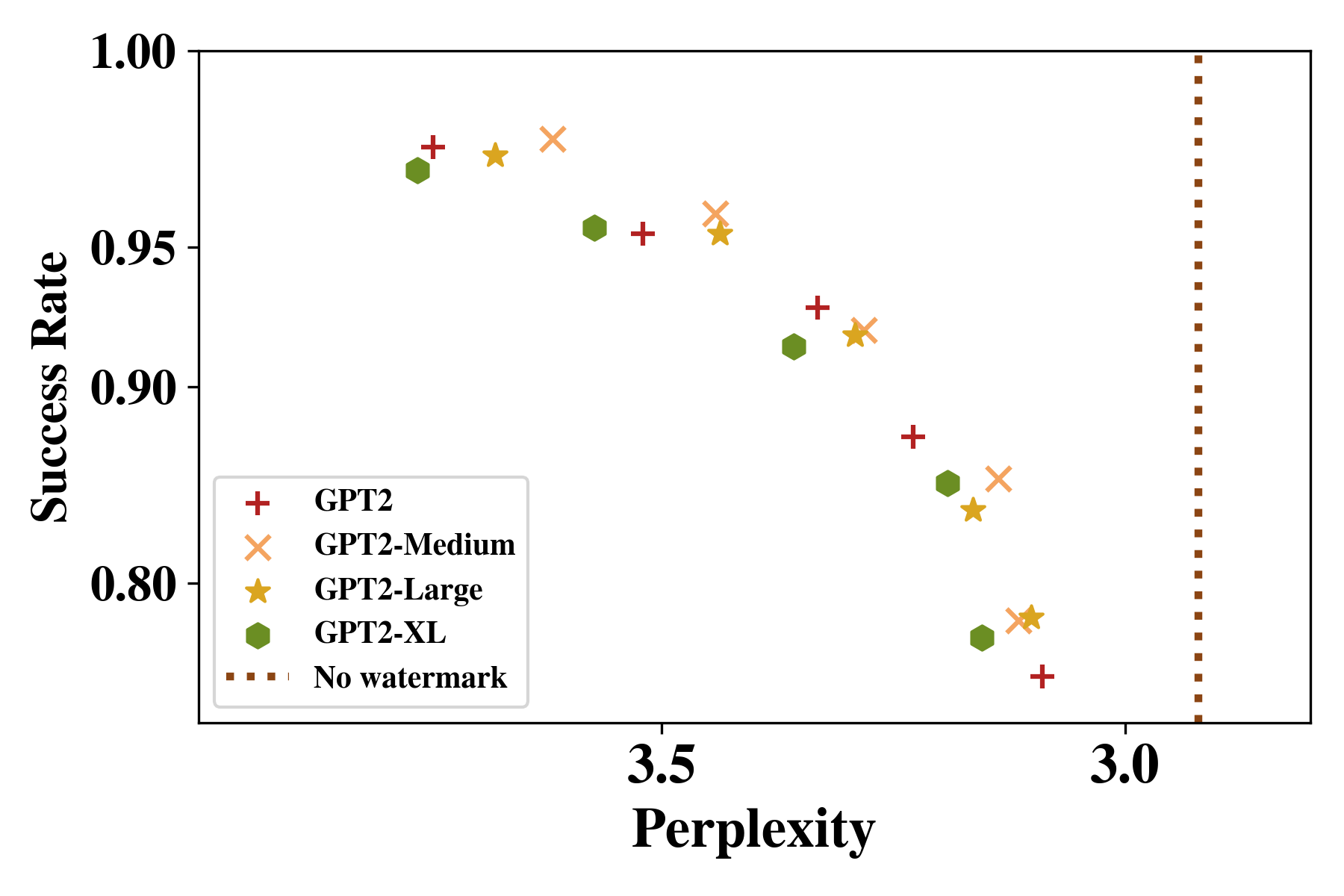}
    \caption{Illustration of $LM_{proxy}$'s impact on watermarking. Larger $LM_{proxy}$ tends to exhibit improved performance.}
    \label{fig: ablation_figs}
\end{figure}

% In Balance-Marking, we use $P_{LM_{proxy}}(v|x_{l-L_{prefix}-1:(l-1)})$ to estimate $P_LLM(v|x_{:(l-1)})$. It seems plausible that using a better $LM_{proxy}$ can provide more precise estimation of $P_LLM(v|x_{:(l-1)})$, thus resulting in better performance. Experiments detailed in Figure~\ref{fig: ablation_figs} verify the hypothesis. Interestingly, GPT2-Large has better performance than GPT2-XL, indicating that such promotion may have an upper limit.
In our Balance-Marking method, we employ $LM_{proxy}$ to generate an estimation for $P_{LLM}(v|\mathbf{t}_{:(l-1)})$. It can be reasonably hypothesized that an enhanced $LM_{proxy}$ could offer a more accurate estimation of $P_{LLM}(v|\mathbf{t}_{:(l-1)})$, thereby leading to superior performance. Experiments detailed in Figure~\ref{fig: ablation_figs} support this hypothesis.\footnote{Here, we run experiments with $\delta\in\{1.0, 1.2, 1.5, 2.0, 3.0\}$. Appendix~\labelcref{sec: The effect of $A$,sec: The impact of L_prefix,sec: The impact of |M|} also use the same $\delta$s.} However, GPT2-Large illustrates better performance in comparison to GPT2-XL, suggesting that the improvement might have a certain ceiling.

% Regrettably, the computational cost also rises with a larger $LM_{proxy}$, as shown in Table~\ref{tab:lm_pub_efficieny}. In practical usage, one may need to balance between time cost and watermark quality.
Unfortunately, as the size of $LM_{proxy}$ increases, there follows a consequential surge in computational cost, clearly demonstrated in Table~\ref{tab:lm_pub_efficieny}. For practical applications, it becomes necessary to establish a balance between time expenditure and the quality of the watermark.

% \begin{table}[t]
%     \centering
%     \begin{tabular}{c|c|c|c|c}
%         \toprule
%         $\boldsymbol{LM_{proxy}}$ & \textbf{GPT2~(124M)} & \textbf{GPT2-Medium~(355M)} & \textbf{GPT2-Larg~(774M)} & \textbf{GPT2-XL~(1.5B)} \\
%         \midrule
%         Encoding Time & 9.50 & 11.69 & 14.08 & 16.38 \\
%         \midrule
%         Decoding Time & 2.21 & 2.37 & 2.57 & 2.83 \\
%         \bottomrule
%     \end{tabular}
%     \caption{The increase in $LM_{proxy}$ size escalates the computational cost during both encoding and decoding.}
%     \label{tab:lm_pub_efficieny}
% \end{table}
\begin{table}[t]
    \centering
    \caption{Growing $LM_{proxy}$ size amplifies the computational cost during both encoding and decoding processes.}
    \begin{tabular}{c|c|c|c|c}
        \toprule
         $\boldsymbol{LM_{proxy}}$ & \makecell{\textbf{GPT2}\\(124M)} & \makecell{\textbf{GPT2-Medium}\\(355M)} & \makecell{\textbf{GPT2-Large}\\(774M)} & \makecell{\textbf{GPT2-XL}\\(1.5B)} \\
        \midrule
        Encoding Time~(s) & 9.50 & 11.69 & 14.08 & 16.38 \\
        \midrule
        Decoding Time~(s) & 2.97 & 3.09 & 3.43 & 3.53 \\
        \bottomrule
    \end{tabular}
    \label{tab:lm_pub_efficieny}
\end{table}

\paragraph{Special Case: $LM_{proxy}=LLM$}
\label{app:anothertable}
Table~\ref{tab:lm_pub_efficieny2} and Table~\ref{tab:lm_pub_qua3} show the efficiency and quality of different $LM_{proxy}$s and the $LM_{proxy}=LLM$ case. $LM_{proxy}=LLM$ has the best quality, but is slower than $LM_{proxy}=\textrm{GPT2}$.

\begin{table}[t]
    \centering
    \caption{Efficiency of different $LM_{proxy}$s and the $LM_{proxy}=LLM$ case.}
    \begin{tabular}{c|c|c|c|c|c}
        \toprule
         $\boldsymbol{LM_{proxy}}$ & \makecell{\textbf{GPT2}\\(124M)} & \makecell{\textbf{GPT2-Medium}\\(355M)} & \makecell{\textbf{GPT2-Large}\\(774M)} & \makecell{\textbf{GPT2-XL}\\(1.5B)} & \makecell{$\mathbf{LLM}$ \textbf{self}\\(OPT-1.3B)}\\
        \midrule
        Encoding Time~(s) & 9.50 & 11.69 & 14.08 & 16.38 & 11.05 \\
        \midrule
        Decoding Time~(s) & 2.97 & 3.09 & 3.43 & 3.53 & 4.07\\
        \bottomrule
    \end{tabular}
    \label{tab:lm_pub_efficieny2}
\end{table}

\begin{table}[t]
    \centering
    \caption{The watermark success rate of different $LM_{proxy}$s and the $LM_{proxy}=LLM$ case under different PPLs.}
    \begin{tabular}{c|c|c|c}
        \toprule
        %  $\boldsymbol{PPL}$ & 3.2 & 3.35 &
        %  \makecell{\textbf{GPT2}\\(124M)} & \makecell{\textbf{GPT2-Medium}\\(355M)} & \makecell{\textbf{GPT2-Large}\\(774M)} & \makecell{\textbf{GPT2-XL}\\(1.5B)} & \makecell{$\mathbf{LLM}$ \textbf{self}\\(OPT-1.3B)}\\
        % \midrule
        % Encoding Time~(s) & 9.50 & 11.69 & 14.08 & 16.38 & 11.05 \\
        % \midrule
        % Decoding Time~(s) & 2.97 & 3.09 & 3.43 & 3.53 & 4.07\\
% \vspace{2pt}    
\textbf{PPL} & \textbf{3.2} & \textbf{3.35} & \textbf{3.5} \\ \hline
 \makecell{\textbf{GPT2} (124M)} & 83.1 & 91.3 & 95.3 \\
\makecell{\textbf{GPT2-Medium} (355M)} & 84.9 & 93.2 & 95.1 \\
\makecell{\textbf{GPT2-Large} (774M)} & 88.6 & 93.8 & 96.7 \\
\makecell{\textbf{GPT2-XL} (1.5B)} & 86.4 & 93.4 & 96.0 \\
\makecell{$\mathbf{LLM}$ \textbf{self} (OPT-1.3B)} & 91.7 & 94.8 & 96.9 \\ 
        \bottomrule
    \end{tabular}
    \label{tab:lm_pub_qua3}
\end{table}

\subsection{The impact of the pre-mapping space size $A$}
\label{sec: The effect of $A$}
% The hyper-parameter $A$ introduced in $LM_{proxy}$ aims to reduce the number of $V_{m,\mathbf{t}_{:(l-1)}}$s that need to be computed. A smaller $A$ means lower computation costs during decoding, as shown in Figure~\ref{tab:A_efficieny} (encoding time is roughly the same, since we only needs to calculate one $V_{m,\mathbf{t}_{:(l-1)}}$ when encoding the message $m$ into text).
The purpose of introducing hyper-parameter $A$ in Balance-Marking is to reduce the number of $V_{m,\mathbf{t}_{:(l-1)}}$ to be computed. A lower value of $A$ results in lower computational expenses during decoding, as shown in Table~\ref{tab:A_efficieny}. On the other hand, the cost of encoding remains relatively constant because only a single message $m$ requires the computation of $V_{m,\mathbf{t}_{:(l-1)}}$ during encoding.

\begin{table}[t]
    \centering
    \caption{Investigation into the impact of $A$ on watermark efficiency. A larger $A$ value generally results in a longer decoding time.}
    \begin{tabular}{c|c|c|c|c|c}
        \toprule
        $\boldsymbol{A}$ & \textbf{25} & \textbf{50} & \textbf{100} & \textbf{150} & \textbf{250} \\
        \midrule
        Encoding Time~(s) & 9.39 & 9.48 & 9.50 & 9.22 & 9.37 \\
        \midrule
        Decoding Time~(s) & 1.89 & 2.27 & 2.97 & 4.49 & 6.24 \\
        \bottomrule
    \end{tabular}
    \label{tab:A_efficieny}
\end{table}

\begin{figure}[t]
    \centering
    \includegraphics[width=0.8\textwidth]{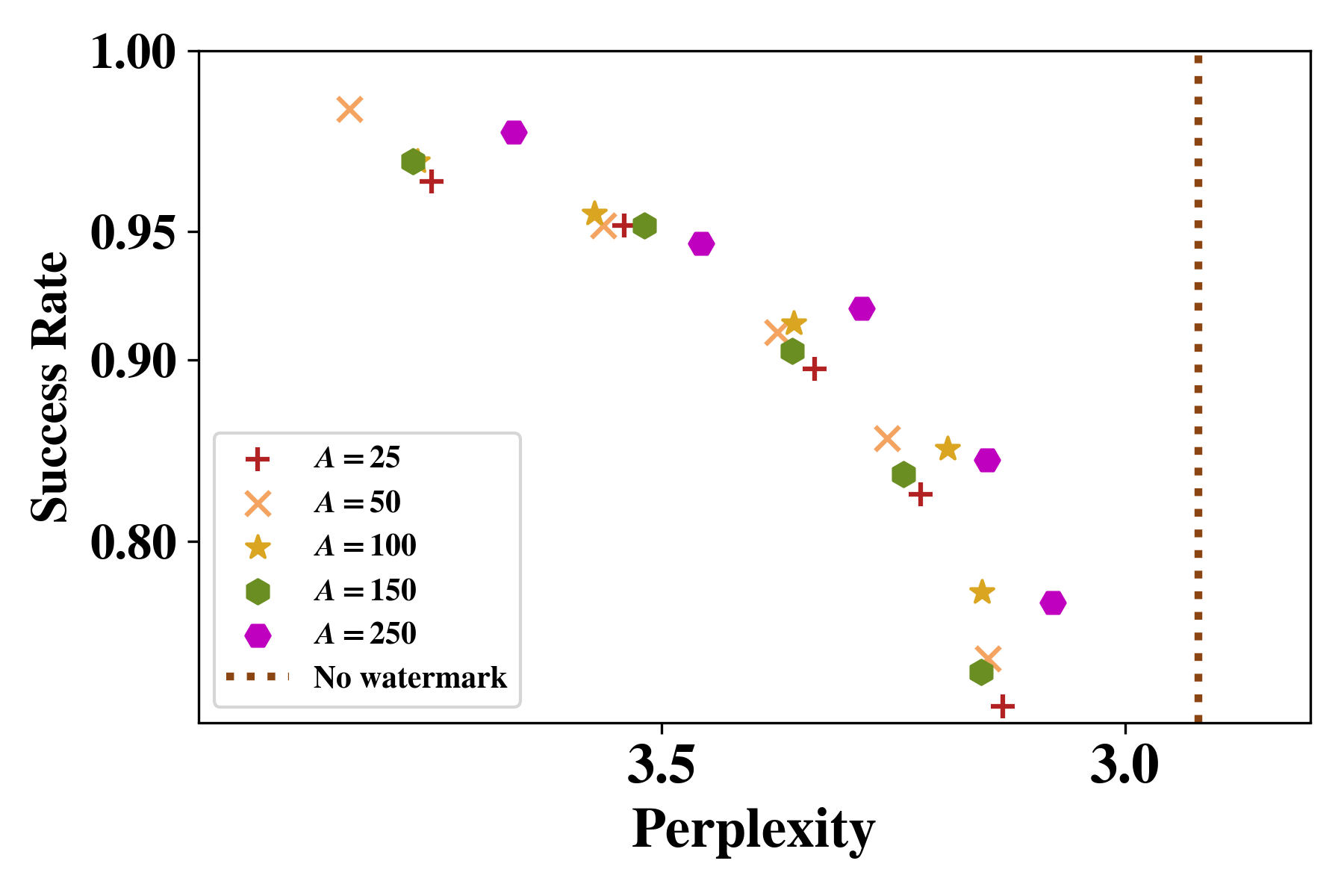}
    \caption{Illustration of $A$'s impact on watermarking quality. A higher $A$ tends to have better watermark quality.}
    \label{fig: ablation_A}
\end{figure}

To explore the impact of $A$ on watermarking quality, we conducted additional experiments as demonstrated in Figure~\ref{fig: ablation_A}. Although a larger $A$ generally leads to enhanced performance, there exist notable exceptions, such as when $A=150$, where the performance is lower than when $A=100$.

To strike a balance between performance and computational efficiency, we elected to use $A=100$ for our main experiments in Section~\ref{sec: main_results}.

\begin{figure}[t]
    \centering
    \includegraphics[width=0.8\textwidth]{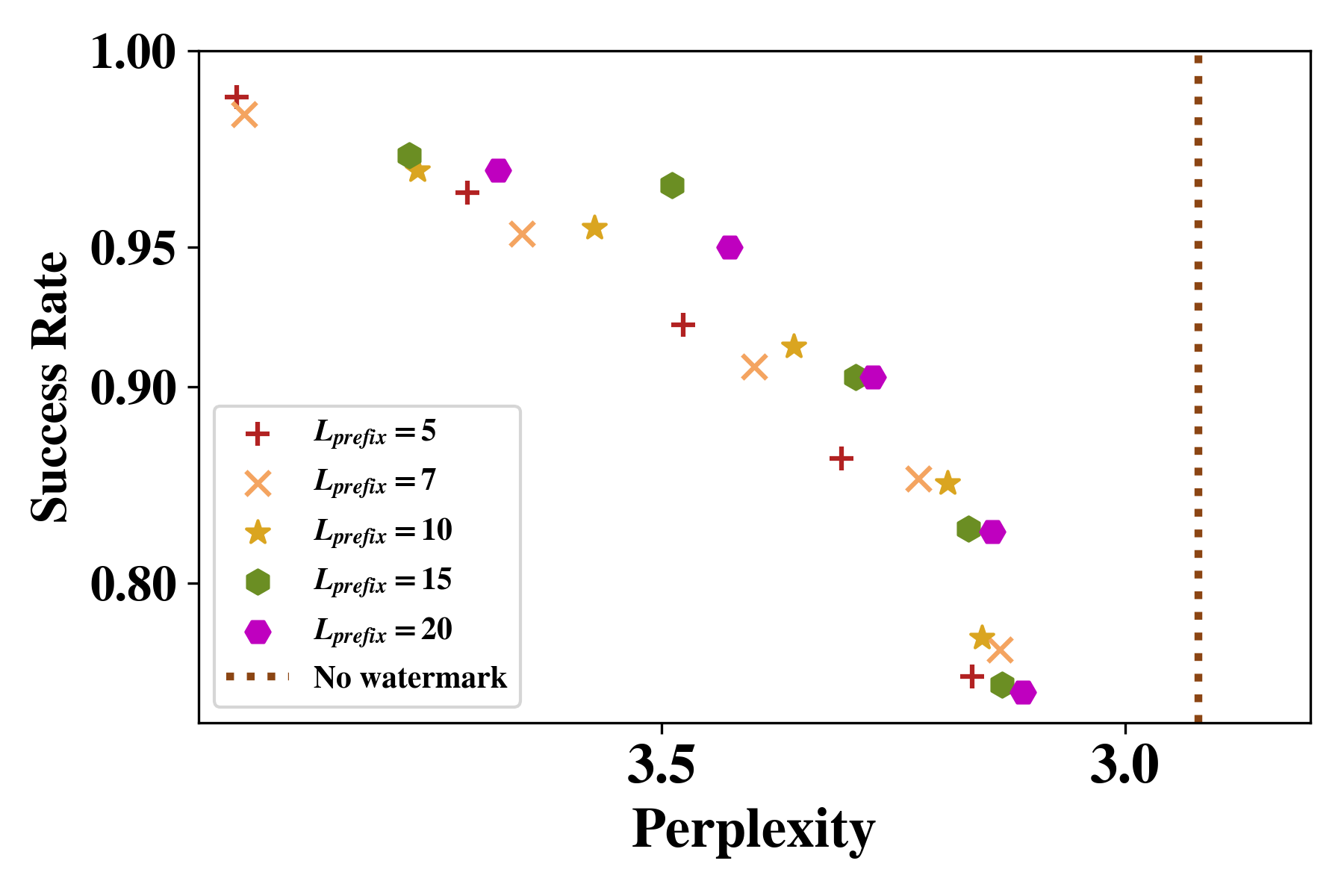}
    \caption{Illustration of $L_{prefix}$'s impact on watermark quality. A too-low $L_{prefix}$ value can degrade the watermark quality.}
    \label{fig: prefix}
\end{figure}

\begin{table}[t]
    \centering
    \caption{Investigation into the impact of $L_{prefix}$ on watermark efficiency. $L_{prefix}$ has a modest impact on watermark efficiency.}
    \begin{tabular}{c|c|c|c|c|c}
        \toprule
        $\boldsymbol{L_{prefix}}$ & \textbf{5} & \textbf{7} & \textbf{10} & \textbf{15} & \textbf{20} \\
        \midrule
        Encoding Time~(s) & 9.52 & 9.03 & 9.50 & 9.29 & 9.10 \\
        \midrule
        Decoding Time~(s) & 3.01 & 3.05 & 2.97 & 2.91 & 2.86 \\
        \bottomrule
    \end{tabular}
    \label{tab: prefix_efficieny}
\end{table}

\begin{table}[t]
    \centering
    \caption{Investigation into the impact of $|\mathcal{M}|$ on watermark efficiency. The size of $\mathcal{M}$ has a modest impact on watermark efficiency.}
    \begin{tabular}{c|c|c|c|c}
        \toprule
        $\boldsymbol{|\mathcal{M}|}$ & $\boldsymbol{2^5}$ & $\boldsymbol{2^7}$ & $\boldsymbol{2^{10}}$ & $\boldsymbol{2^{20}}$ \\
        \midrule
        Encoding Time~(s) & 9.89 & 10.02 & 9.83 & 10.05 \\
        \midrule
        Decoding Time~(s) & 2.37 & 2.34 & 2.36 & 3.13 \\
        \bottomrule
    \end{tabular}
    \label{tab:M_efficieny}
\end{table}

\begin{figure}[t]
    \centering
    \includegraphics[width=0.8\textwidth]{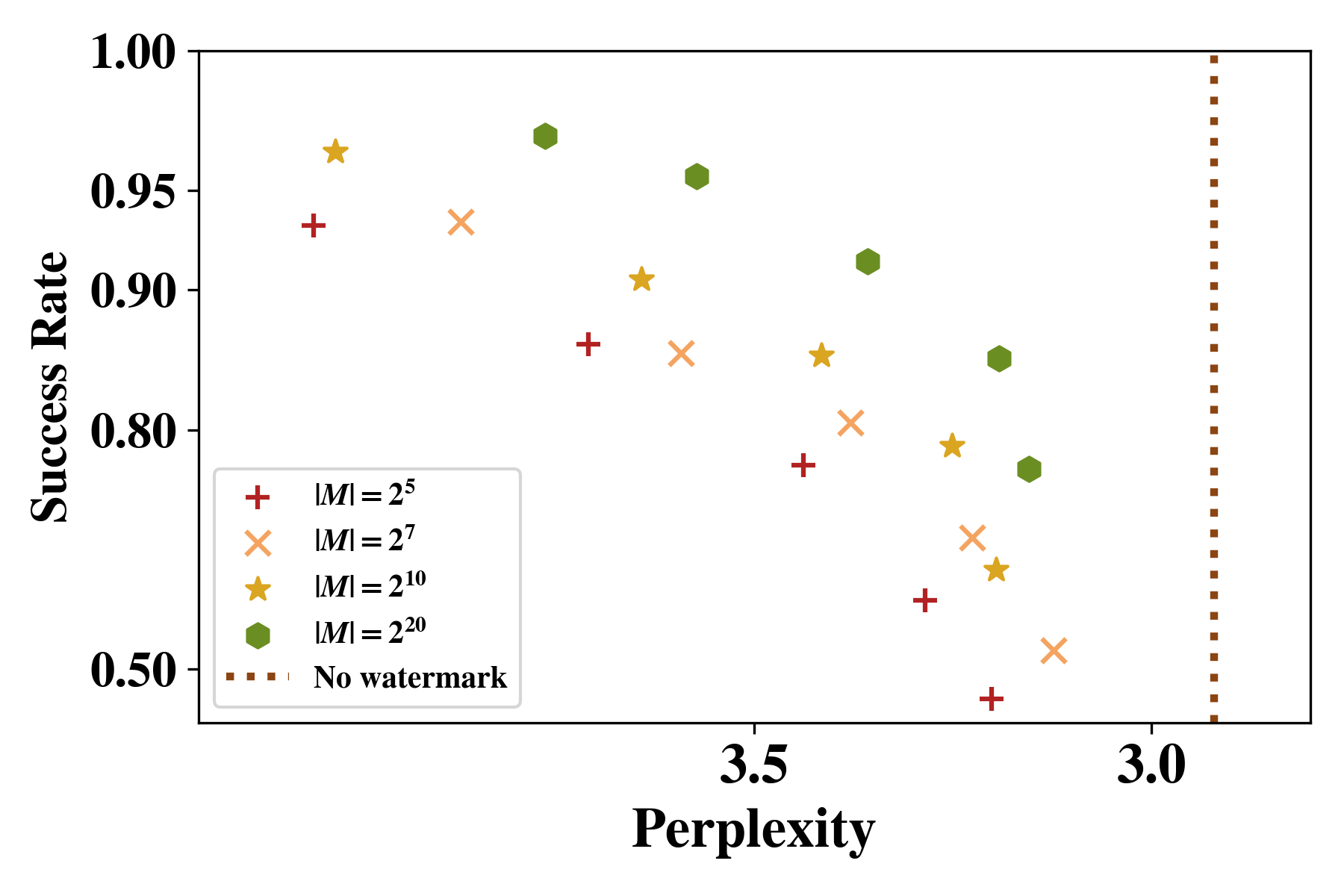}
    \caption{Illustration of $|\mathcal{M}|$'s impact on watermark quality. A larger $\mathcal{M}$ results in better watermark quality.}
    \label{fig: ablation_message_code_len}
\end{figure}

\subsection{The impact of the truncation length $L_{prefix}$}
\label{sec: The impact of L_prefix}
The parameter $L_{prefix}$, which is part of Eq.~(\ref{eq: V_m condition proxy}), impacts the quality of the watermark in two significant ways. (1) A longer $\mathbf{t}_{(l-1- L_{prefix}):(l-1)}$ in $P_{LM_{proxy}}(v| \mathbf{t}_{(l-1- L_{prefix}):(l-1)})$ can potentially provide a superior approximation of $P_{LLM}(v|\mathbf x^{prompt},\mathbf{t}_{:(l-1)})$. However, (2) an increased $\mathbf{t}_{(l-1- L_{prefix}):(l-1)}$ length will reduce the number of effective tokens in $\mathbf t$ available for encoding and decoding. This reduction occurs because  $\mathbf t_{:L_{prefix}}$ lacks sufficient preceding words to form a $\mathbf{t}_{(l-1- L_{prefix}):(l-1)}$, which consequently results in their exclusion from encoding and decoding. Figure~\ref{fig: prefix} illustrates the influence of $L_{prefix}$. Generally speaking, a relatively low $L_{prefix}$ value can degrade the watermark quality, while a moderate $L_{prefix}$ like 10 achieves similar performance to larger values such as 15 or 20.

The time cost for encoding and decoding remains relatively consistent across varying $L_{prefix}$ values, as demonstrated in Table~\ref{tab: prefix_efficieny}. Interestingly, a larger $L_{prefix}$ might lead to a decrease in time cost, aligning with the aforementioned explanation that a more extended $L_{prefix}$ results in a shorter effective $\mathbf t$ utilized in the encoding and decoding stages.

\subsection{The impact of the message space size $|\mathcal{M}|$}
\label{sec: The impact of |M|}
% With a fixed payload information coding rate, a larger size of $|\mathcal{M}|$ means more tokens for a single message. Consider the situation that parts of the text are low-entropy areas where watermark is difficult to apply, with more tokens, it is unlikely that all text for a message is of low-entropy, thus providing high-entropy parts for the message to be embedded. Also, suppose $|\mathcal{M}|$ is small, then a piece of 20-bit information may be split to 4 pieces of 5-bit information to be encoded respectively. If any of 4 pieces of 5-bit information fails to be encoded, the whole 20-bit information fails to be encoded. In other words, the error may be accumulated to a high degree. Figure~\ref{fig: ablation_message_code_len} verifies that a larger $\mathcal{M}$ corresponds to better watermark quality. Here, we ask the model to generate 210 tokens, since for $|\mathcal{M}| = 2^7$ and a coding rate of 7 token per bit, one message corresponds to 70 tokens.
Increasing the size of $|\mathcal{M}|$ under a fixed payload information coding rate results in a larger number of tokens available for embedding a single message. In text watermarking scenarios, certain parts of the text often exhibit low entropy, thereby challenging the watermark encoding process~\citep{watermark_llm}. However, an increase in the number of tokens mitigates this problem, as it improves the likelihood of having high-entropy sections where the message can be effectively embedded. Moreover, let's consider an example where $|\mathcal{M}|$ is relatively small. In such a case, a piece of information of 20 bits might need to be divided into four separate chunks of 5 bits each to be encoded. If encoding fails for any of these 5-bit information segments, the encoding of the entire 20-bit information fails. This phenomenon potentially results in a significant accumulation of errors. 

Figure~\ref{fig: ablation_message_code_len} validates the analysis above by demonstrating a strong correlation between larger $|\mathcal{M}|$ and better watermark quality. Here, the experiment utilizes the $LLM$ to generate 210 tokens for encoding and decoding, since when $|\mathcal{M}| = 2^7$ and the coding rate is 10 tokens per bit, one message corresponds to 70 tokens.

% Interstingly, the encoding and decoding time are similar when $|M|$ increases. This indicates the calculation of $P_{LM_{proxy}}(v|x_{(l-1-L_{prefix}):(l-1)}$ and $V_{m,\mathbf{t}_{:(l-1)}}$ takes up most of the time (note here the number of $V_{m,\mathbf{t}_{:(l-1)}}$ needing computation depends on $A$ instead of $m$). However, $|\mathcal{M}|$ can not be enlarged limitlessly. Suppose $|\mathcal{M}|$ is $2^{40}$, it is nearly impossible to calculate $P_w(\mathbf t| m')$ in Eq.~(\ref{eq: decoding messages}) for all messages $m'\in\mathcal{M}$ during decoding. 
Table~\ref{tab:M_efficieny} shows the time costs of the encoding and decoding processes when $|\mathcal{M}|$ increases. The time expenditure for encoding remains relatively stable, since there is only one message for which the calculation of $P_w$ is necessary. Contrastingly, the decoding process experiences a noticeable increase in time cost when $|\mathcal{M}|$ rises up to $2^{20}$. Thus, while an increase in $|\mathcal{M}|$ steadily improves watermark quality, the size of $\mathcal{M}$ can not be increased unlimitedly.\footnote{For instance, if $|\mathcal{M}|$ equals $2^{40}$, it becomes exceedingly difficult to calculate $P_w(\mathbf t| m')$ in Eq.~(\ref{eq: decoding messages}) for all messages $m'\in\mathcal{M}$ during the decoding process.}

\subsection{Hash Scheme} 
\label{appsec: hash}
Following the hash implementation in~\citet{watermark_llm}, in the case of both the Vanilla-Marking and Balance-Marking methods, we make use of the last token in $\mathbf{t}_{:(l-1)}$ to calculate $h(v,m,\mathbf{t}_{:(l-1)})$ in Eq.~(\ref{eq: vanilla_CTWL raw}) and $h(m,\mathbf{t}_{:(l-1)})$ in Algorithm~\ref{algo: choose V_{m,x_{:(l-1)}}_old}, i.e. $h(v,m,\mathbf{t}_{:(l-1)})=h(v,m,t_{l-1})$ and $h(m,\mathbf{t}_{:(l-1)})=h(m,t_{l-1})$.

\section{Implementation of Substitution Attacks}
\label{appendix: details_sub_attack}
For a chosen text to be tested on, we arbitrarily pick an unaltered token each time. This token is then masked and the model is asked to predict it. In the event that a token's predicted logit surpasses the predicted logit of the original token minus 1.0, we replace the original token with the new one. With a designated substitution ratio $\alpha$ and a sentence consisting of $L$ tokens, we continue this process until the substituted tokens reach the value of $\alpha L$ or after $3\alpha L$ attempts are made. From empirical analysis, we establish that such replacements result in a marginal increase in PPL by around 0.1.

\section{Substitution Attacks with a substitution ratio of 10\%}
\label{appendix:sub-10}

\begin{figure}[t]
    \centering
    \begin{subfigure}[b]{0.48\textwidth}
        \includegraphics[width=\linewidth]{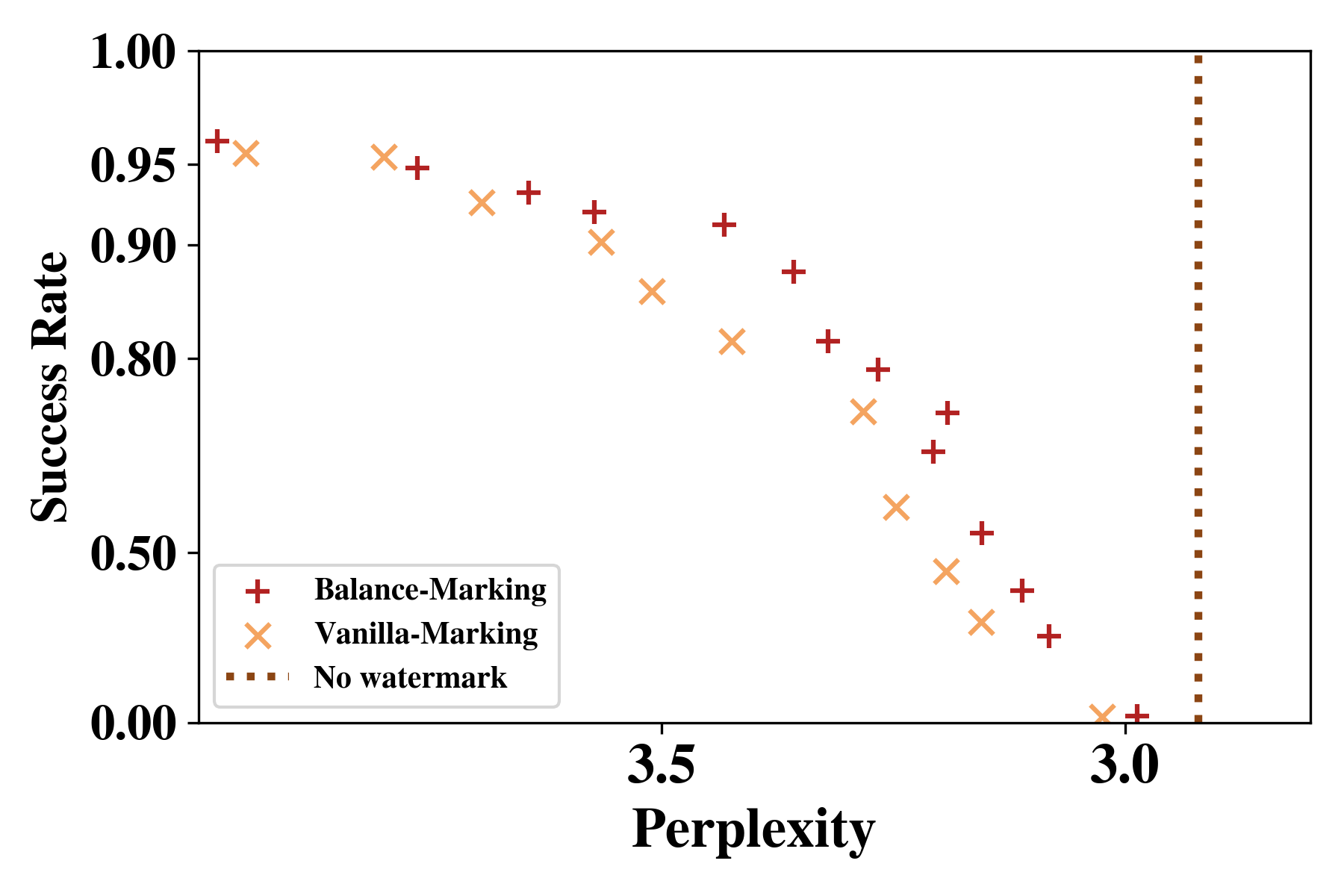}
        \caption{Substitution Attacks (substitution ratio = 5\%).}
        \label{fig:app_sub-0.05}
    \end{subfigure}
    \hfill
    \begin{subfigure}[b]{0.48\textwidth}  
        \includegraphics[width=\linewidth]{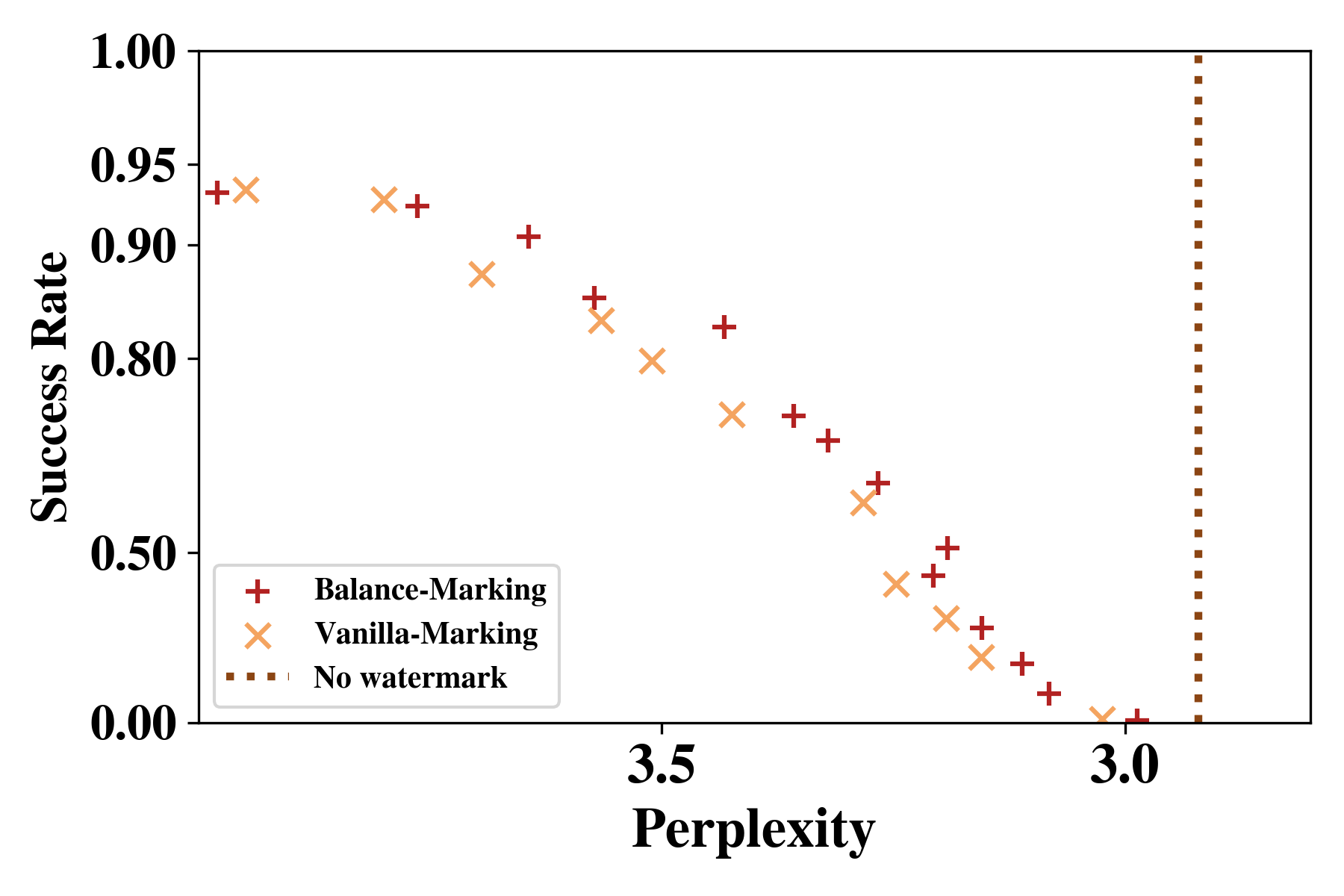}
        \caption{Substitution Attacks (substitution ratio = 10\%).}
        \label{fig:app_sub-0.1}
    \end{subfigure}
\caption{The relationship between the success rate after Copy-Paste Attacks / Substitution Attacks and the PPL. Balance-Marking outperforms Vanilla-Marking under both attacks.}
\label{fig:sub_attacks}
\end{figure}

Figure~\ref{fig:sub_attacks} compares the results under a substitution ratio of 5\% and 10\%. A higher substitution results in a lower success rate, and the performance of Balance-Marking and Vanilla-Marking becomes closer from Figure~\ref{fig:app_sub-0.05} to Figure~\ref{fig:app_sub-0.1}, indicating that Substitution Attacks may do more hurt to Balance-Marking than Vanilla-Marking. This can be attributed to the fact that $P_w$ of Balance-Marking relies on $\mathbf{t}_{(l-1-L_{prefix}):(l-1)}$~(see Eq.~(\ref{eq: V_m condition proxy})), while $P_w$ of Vanilla-Marking only depends on $t_{(l-1)}$, since we only use the last token of $\mathbf{t}_{:(l-1)}$ to calculate $h(m,\mathbf{t}_{:(l-1)})$ (see Section~\ref{subsec: experimental settings}). So, Balance-Marking will be affected more when more tokens are substituted.

\section{Results of LLaMA-7B and LLaMA-13B at a coding rate of 5}
\label{appendix: scaling}
\begin{figure}[t]
    \centering
\includegraphics[width=0.8\textwidth]{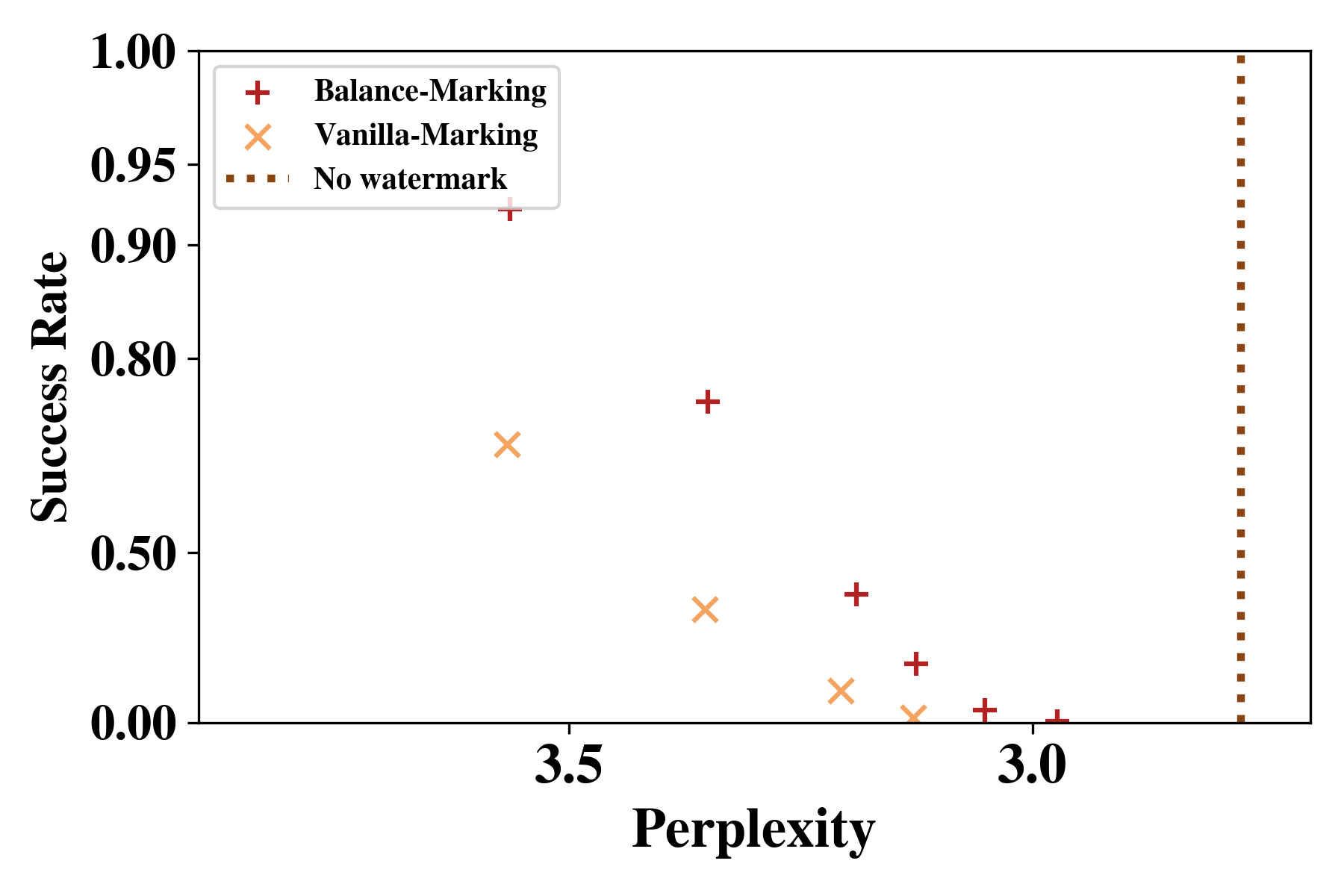}
    \caption{Balance-Marking outperforms Vanilla-Marking under LLaMA-7B and coding rate of 5 tokens per bit).}
    \label{fig: scaling-7b-5}
\end{figure}

\begin{figure}[t]
    \centering
\includegraphics[width=0.8\textwidth]{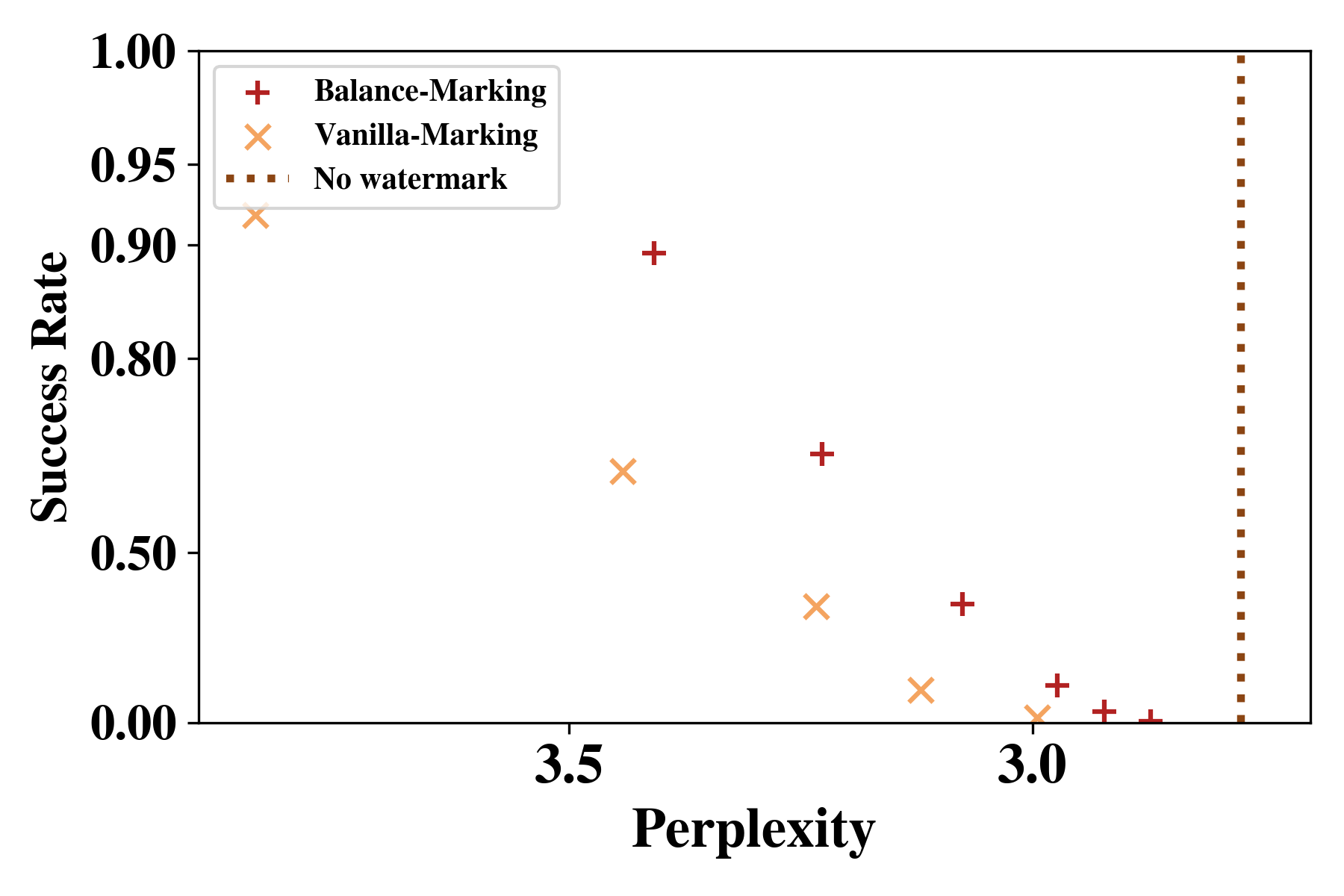}
    \caption{Balance-Marking outperforms Vanilla-Marking under LLaMA-13B and coding rate of 5 tokens per bit).}
    \label{fig: scaling-13b-5}
\end{figure}

Besides experiments with a coding rate of 10 in Section~\ref{para:scaling}, we also tested the coding rate of 5. The results are shown in Figure~\labelcref{fig: scaling-13b-5,fig: scaling-7b-5}. Balance-Marking outperforms Vanilla-Marking in all cases.

\section{Illustrative Examples of Watermarked Texts}
\label{sec: examples}
In this section, we present some examples generated using our watermarking approach (Table~\ref{examples-finalpart}). These examples serve as a case study to demonstrate the quality of our watermarking technique. All the generated sentences are truncated at 200 tokens.

\begin{table}[h]
\centering
\caption{Examples of generated texts.}
\begin{tabular}{|l|l|}
\hline
\textbf{Case Study}                                              & \textbf{Text}                                                    \\ \hline
\parbox{3cm}{Prompt}                                               & \parbox{10cm}{\vspace{2pt}An EgyptAir jet sent a series of signals indicating that smoke had been detected on board before it crashed into the Mediterranean on Thursday, France’s air accident investigation agency said on Saturday. A spokesman for the BEA agency said the signals did not indicate what caused the smoke or fire on board the plane, which plunged into the sea with 66 people on board as it was heading from Paris to Cairo. But they offered the first clues as to what unfolded in the moments before the crash. One aviation source said that a fire on board would likely have generated multiple warning signals, while a sudden explosion may not have generated any. The army spokesman published pictures on Saturday on its official Facebook page of the recovered items, which included blue debris with EgyptAir markings, seat fabric with designs in the airline’s colours, and a yellow lifejacket. It was the third blow since October to Egypt’s travel industry, still reeling from political unrest following the 2011 uprising that ousted Hosni Mubarak. A suspected Islamic State bombing brought down a Russian airliner after it took off from Sharm el-Sheikh airport in late October, killing all 224 people on board, and an EgyptAir plane was hijacked in March by a man wearing a fake suicide belt. “These messages do not allow in any way to say what may have caused smoke or fire on board the aircraft,” said a spokesman for the French BEA agency, which is assisting an official Egyptian\vspace{2pt}}       \\ \hline

\parbox{3cm}{No-watermark\\(OPT-1.3B)}                                         & \parbox{10cm}{\vspace{2pt}inquiry into the crash. “We will continue our work until we are able to give you more information,” he added. Egyptian authorities have so far given few details about the crash, saying only that it happened shortly after take-off at 4:18 p.m. (1318 GMT) on Thursday. The flight data recorder and cockpit voice recorder were recovered from the ocean floor on Friday. The black box recorders are expected to be handed over to investigators later this month. BEA spokesman Jean-Francois Delfraissy told a news conference on Saturday that the black boxes contained “a great deal of information”. He declined to comment on whether there were any indications of mechanical failure. He also said the BEA had received no reports of any distress calls from the plane before it disappeared from radar screens. French President Francois Hollande said on Saturday that his thoughts were with the families of those who died in the crash.\vspace{2pt}}                                                          \\ \hline

\parbox{3cm}{Vanilla-Marking\\(OPT-1.3B)\\(watermark success rate = 93\%)} & \parbox{10cm}{\vspace{2pt}probe. “We will continue our work until we can be certain about the cause.” Egyptian officials said earlier on Saturday that they had found no evidence of foul play in the crash. They also said there were no signs of mechanical failure on the Airbus A320, which was flying at 31,000 feet (9,800 meters) when it disappeared from radar screens shortly after take-off from Oran, near the Libyan border, at 2:24 p.m. local time (0824 GMT). The wreckage was found some 20 kilometres (12.5 miles) off the Libyan coast on Saturday afternoon. It is unclear how long it has been in the water or how deep it is. The search is being conducted by a naval patrol vessel, two ships from the Italian navy, one from the United Arab Emirates, one from Greece and a helicopter from the French Navy. The aircraft was carrying 149 passengers and nine crew members, including 82 Egyptians, 12 Italians,\vspace{2pt}} 
\\ \hline
\end{tabular}
\label{examples-finalpart}
\end{table}

\begin{table}[h]
\centering
\caption*{Table \thetable\ (Continued): Examples of generated texts.}
\begin{tabular}{|l|l|}
\hline
\textbf{Case Study}                                              & \textbf{Text}                                                    \\ \hline

\parbox{3cm}{Balance-Marking\\(OPT-1.3B)\\(watermark success rate = 95\%) }       & \parbox{10cm}{\vspace{2pt}inquiry into the crash. “They are just indications that there could have been some kind of activity on the part of the aircraft,” he added. He declined to comment further. French President Francois Hollande, who visited the crash site in the southern city of Ismailia with Prime Minister Manuel Valls on Saturday, has called for an international investigation into the crash. Egypt’s civil aviation authority said in a statement on Saturday that an initial examination of the black box flight data recorder showed that “there were no signs of a mechanical problem with the aircraft.” It added: “The preliminary findings of the investigation suggest that there was a technical fault that led to the loss of communication between the aircraft and the ground control centre.” CAUGHT BY SURVIVORS Survivors described hearing a loud noise and then seeing flames coming out of the rear of the Airbus A320, one of the world’s most modern planes\vspace{2pt}}                                                         \\ \hline
\parbox{3cm}{No-watermark\\(LLaMA-7B)}       & \parbox{10cm}{\vspace{2pt}aring a fake suicide belt who demanded to be flown to Cyprus. Egyptian President Abdel Fattah al-Sisi declared three days of national mourning and ordered flags to fly at half-mast until Tuesday. “We will spare no effort to find out the reasons behind this disaster,” Prime Minister Sherif Ismail told reporters in Cairo. France’s foreign minister Jean-Marc Ayrault said he had spoken to his Egyptian counterpart Sameh Shoukry and expressed France’s “solidarity” with the families of the victims. French President Francois Hollande is due to speak to Sisi later on Saturday, according to the Elysee presidential palace. German Chancellor Angela Merkel said she was “deeply shaken” by the news of the crash, adding: “Our thoughts are with the families\vspace{2pt}}                                                         \\ \hline
\parbox{3cm}{Vanilla-Marking\\(LLaMA-7B)\\(watermark success rate = 96.8\%)}       & \parbox{10cm}{\vspace{2pt}aring a fake suicide belt who forced it to land in Cyprus. Egyptian officials say there is no evidence so far of a terrorist attack but investigators are treating the incident as a “criminal act”. France’s foreign minister Jean-Marc Ayrault, speaking on BFM TV, said he could not rule out a terrorist attack. “I can’t exclude anything at this stage … We must remain very vigilant,” he said, adding that French experts were taking part in the investigation. Egyptian security officials told Reuters on Friday that a technical glitch was being investigated as the possible cause of the crash. Another senior Egyptian security official, speaking on condition of anonymity, said investigators had found no trace of explosives or traces of chemicals used to make explosive devices on the remains of the Airbus A320. Investigators also found\vspace{2pt}}                                                         \\ \hline
\parbox{3cm}{Balance-Marking\\(LLaMA-7B)\\(watermark success rate = 96.2\%)}       & \parbox{10cm}{\vspace{2pt}aring a fake suicide belt who demanded to be flown to Cyprus. He was overpowered and the plane landed safely in Cyprus. British Foreign Secretary Philip Hammond said he had spoken to Egyptian President Abdel Fattah al-Sisi about the incident. “I expressed our condolences to the families of the victims,” he told reporters in Brussels. Hammond declined to comment when asked whether Britain would review security arrangements at British airports, including London’s Heathrow, which has direct flights to Egypt’s main tourist destinations. The United Nations Security Council condemned the “outrageous and cowardly terrorist attack” but made no mention of possible retaliatory action against militants in Sinai. The council also urged all states to cooperate with the Egyptian authorities in their efforts to bring perpetrators to justice.\vspace{2pt}}                                                         \\ \hline
\end{tabular}
\end{table}

\begin{table}[h]
\centering
\caption*{Table \thetable\ (Continued): Examples of generated texts.}
\begin{tabular}{|l|l|}
\hline
\textbf{Case Study}                                              & \textbf{Text}                                                    \\ \hline

\parbox{3cm}{No-watermark\\(LLaMA-13B)}       & \parbox{10cm}{\vspace{2pt}	aring a fake suicide belt. Egyptian Prime Minister Sherif Ismail told reporters at the crash site: “There is no indication so far of any terrorist or sabotage act.” He added that investigators were looking into all possible causes including mechanical failure, human error and weather conditions. “We are working very hard to find the black box,” he said, referring to the cockpit voice recorder and flight data recorder that could help explain why the Airbus A320 plummeted from 37,000 feet (11,280 metres) into the sea. French President Francois Hollande, whose country lost 54 citizens in the disaster, has ordered an inquiry to be carried out jointly with Egyptian authorities. The BEA will lead the French side of the probe. The BEA spokesman said the search for the wreckage\vspace{2pt}}                                                         \\ \hline
\parbox{3cm}{Vanilla-Marking\\(LLaMA-13B)\\(watermark success rate = 98\%)}       & \parbox{10cm}{\vspace{2pt}aring a suicide belt who forced it to divert to Cyprus. Egyptian Prime Minister Sherif Ismail told reporters on Saturday that investigations were focusing on the possibility of a “terrorist attack” but stressed this could not yet be confirmed. “There is no clear indication at this time,” he said, adding that Egypt was cooperating with other countries to find out the cause of the crash. Investigators are combing through the wreckage of the Airbus (AIR.PA) A320 found 295 km north of the coastal city of Alexandria, searching for the black box flight recorders that will provide crucial clues to the cause of the crash. France’s Bureau d’Enquetes et d’Analyses pour la Securite du Trafic Aérien (BEA), which is leading the probe into the crash, said\vspace{2pt}}                                                         \\ \hline
\parbox{3cm}{Balance-Marking\\(LLaMA-13B)\\(watermark success rate = 97\%)}       & \parbox{10cm}{\vspace{2pt}aring a fake suicide belt who forced it to divert to Cyprus. He surrendered and was arrested after giving himself up. The cause of Thursday’s crash remains unknown, but the focus has turned to the possibility of a technical failure, terrorism or a deliberate act by the pilot or co-pilot, given their high level of training. The Airbus A320 is a workhorse of worldwide aviation. It has a good safety record, with only two fatal accidents in the past 15 years – one of them the Germanwings disaster in the French Alps last year, when a co-pilot appears to have intentionally crashed the plane, killing all 150 people on board. The other was an A320 operated by Indonesian budget carrier Adam Air that crashed into the sea off the coast of Sulawesi in 2007,\vspace{2pt}}                                                         \\ \hline
\end{tabular}
\end{table}

\end{document}

%% file: math_commands.tex
\usepackage{amsmath,amsfonts,bm}

\def\eqref#1{equation~\ref{#1}}
\def\1{\bm{1}}

\DeclareMathAlphabet{\mathsfit}{\encodingdefault}{\sfdefault}{m}{sl}
\SetMathAlphabet{\mathsfit}{bold}{\encodingdefault}{\sfdefault}{bx}{n}